\documentclass[conference]{IEEEtran}
\IEEEoverridecommandlockouts
\usepackage{cite}
\usepackage{amsmath,amssymb,amsfonts}
\usepackage{algorithmic}
\usepackage{graphicx}
\usepackage{textcomp}
\usepackage{xcolor}

\def\BibTeX{{\rm B\kern-.05em{\sc i\kern-.025em b}\kern-.08em
		T\kern-.1667em\lower.7ex\hbox{E}\kern-.125emX}}
\usepackage{verbatim}
\usepackage[colorlinks,
linkcolor=blue,
anchorcolor=blue,
citecolor=blue]{hyperref}
\usepackage{amsmath,amssymb}
\usepackage{algorithmic,algorithm}

\usepackage{graphicx,textcomp,xcolor}
\usepackage{multirow, tabularx, ragged2e}
\newcolumntype{C}{>{\Centering\arraybackslash}X}

\usepackage{float}

\begin{document}
\title{BERTopic-Driven Stock Market Predictions: Unraveling Sentiment Insights}

\author{\IEEEauthorblockN{1\textsuperscript{st} Enmin Zhu}
\IEEEauthorblockA{\textit{Computer and Information Science} \\
\textit{FST}\\
Macau, China \\
mc35330@umac.mo}}

\maketitle

\begin{abstract}
This paper explores the intersection of Natural Language Processing (NLP) and financial analysis, focusing on the impact of sentiment analysis in stock price prediction. We employ BERTopic, an advanced NLP technique, to analyze the sentiment of topics derived from stock market comments. Our methodology integrates this sentiment analysis with various deep learning models, renowned for their effectiveness in time series and stock prediction tasks. Through comprehensive experiments, we demonstrate that incorporating topic sentiment notably enhances the performance of these models. The results indicate that topics in stock market comments provide implicit, valuable insights into stock market volatility and price trends. This study contributes to the field by showcasing the potential of NLP in enriching financial analysis and opens up avenues for further research into real-time sentiment analysis and the exploration of emotional and contextual aspects of market sentiment. The integration of advanced NLP techniques like BERTopic with traditional financial analysis methods marks a step forward in developing more sophisticated tools for understanding and predicting market behaviors.
\end{abstract}


\section{Introduction}
In traditional finance, the Efficient Market Hypothesis posits that stock prices consistently reflect the rational present value of a firm's expected future cash flows, driven by ``dispassionate'' investors. However, the rapid evolution of stock markets has led to a proliferation of diverse investor types participating in these markets, alongside a growing number of financial anomalies that defy the tenets of the Efficient Market Hypothesis \cite{li2017web}. Numerous studies have highlighted that stock market participants are not entirely rational. Consequently, conventional theoretical models in behavioral finance are not competent sufficiently to comprehensively expalin the influence of investor sentiment on the stock market within the context of the social network environment \cite{wang2018combining}.

The stock market represents a quintessential complex system, comprising numerous stakeholders. In the contemporary landscape, the ebb and flow of the stock market intertwine with myriad factors and different types of information, encompassing various news sources and spanning diverse fields like psychology, politics, and economics \cite{guo2017can}. In an era marked by the rise of online marketing and increased accessibility for ordinary individuals to partake in the financial markets, individual investors have emerged as pivotal players in the multifaceted activities of this domain. The sentiment expressed by these individual investors, a significant cohort within the market, holds an ever-expanding influence on stock market dynamics. As social media continues its development, internet users have grown increasingly inclined to share their perspectives on stocks via online platforms. In many respects, the sentiment conveyed through these online commentaries serves as a genuine reflection of their sentiments towards these stocks. This underscores the profound impact of market sentiment on the irrational oscillations within the stock market, underscoring the compelling necessity of delving into sentiment analysis for the prediction of stock market trends.

Certain researchers have previously employed the LDA-POS model for stock price movement prediction \cite{derakhshan2019sentiment}. Latent Dirichlet Allocation (LDA), while a widely used topic modeling algorithm, relies on the bag-of-words model and does not take into account word order information. However, the efficacy of this approach has room for improvement. In recent times, Natural Language Processing (NLP) technologies have emerged as promising tools for contributing to stock prediction. This paper argues that BERT, with its ability to consider word order and context, is better suited for sentiment analysis in the context of stocks, promising more precise predictions. In order to solve the specific issue about the topic extraction of text, BERTopic \cite{grootendorst2022bertopic} is a state-of-the-art language model, which is a topic modeling algorithm founded on the pre-trained language model BERT (Bidirectional Encoder Representations from Transformers) which is capable to considers the semantics and contextual information of the text. This enables us to computationally analyze extensive volumes of subjective information extracted from online platforms, and to establish a topic vector space for effective topic modeling, thereby enhancing the accuracy of our sentiment analysis in the realm of stock market forecasting. 

BERTopic excels in automatically extracting the themes present in stockholders' comments from textual data and clustering texts that share similar themes. This is achieved by leveraging pre-trained BERT models to generate embeddings of text, transforming text data into continuous vector representations. Subsequently, these embeddings are subjected to clustering by a commonly used algorithm, such as HDBSCAN, which groups similar vectors into the same topic or cluster. Additionally, the Uniform Manifold Approximation and Projection (UMAP) technique is employed to reduce the dimensionality of the embeddings before clustering the documents. It is worth noting that BERTopic is an unsupervised topic modeling approach, eliminating the need for pre-labeled topic labels or training data. Initial experiments applying BERTopic techniques to the analysis of social media text data have yielded promising results \cite{sanchez2022travelers}.

In this paper, we utilize BERTopic model for the analysis of stock comments. Our text dataset comprises stock comments obtained by web crawling various internet websites related to the stock market. Given that the data set is constructed of stock-related comments from platforms such as Reddit or Seeking Alpha without utilizing predefined topic tags, it becomes imperative to employ clustering algorithms to categorize these comments into different topics. To validate the effectiveness of topic features in stock price prediction, we compare the performance of various commonly used AI models in forecasting stock trends, both with and without the application of topic features. Our dataset is partitioned into two subsets: a training dataset and a test dataset. Subsequently, we apply sentiment analysis to determine if these comments convey bullish or bearish sentiments, and we also extract the specific topics from each comment. Finally, we compare these topic-sentiment-based predictions with the actual daily stock prices sourced from Yahoo Finance. This comparison serves as a means to validate the efficacy of our experimental approach.

Our paper makes several significant contributions:
\begin{itemize} 
\item Pioneering BERTopic Application: This paper is the first to employ the BERTopic model for sentiment analysis in the context of the stock market. This novel approach brings the power of BERT-based language models to bear on the analysis of stock market sentiments.
\item Focus on Topic-Level Analysis: Instead of analyzing individual sentences, we center our analysis on the layer of topics. By extracting topics from sentences and subsequently analyzing datasets composed of these topics, we intuitively enhance the efficiency of our approach compared to directly analyzing the sentiment of individual sentences. This shift in focus allows for a more holistic understanding of sentiment trends within stock-related discussions.
\item Evaluation of BERTopic: We conduct a comprehensive evaluation of the BERTopic model's performance with different commonly used deep learning models in the realm of stock prediction. Given the diverse models, the performance of the models which are applied the topic feature is more nuanced and potentially more informative compared to those without. We integrate the topical features extracted by BERTopic into various neural network architectures, including Long Short-Term Memory (LSTM) networks, Convolutional Neural Networks (CNN), and integrated model of CNN and LSTM, to prove the validity of the topic feature. In addition, we use the BERT based model and VADER to generate the sentiment score for the topics respectively, to demonstrate we can get the better performance with whatever sentiment model. Moreover, we employ a BERT-based model and VADER to respectively generate sentiment scores for various topics. This methodology is implemented to illustrate our capability to achieve superior performance metrics across different sentiment analysis models.

\end{itemize}

\section{Literature Review}
\subsection{The application of sentiment analysis in financial field}
In recent years, the utilization of the sentiment index among market participants has gained traction as a valuable tool for predicting stock market trends. Nevertheless, its applications transcend the realm of stock price prediction. F. Z. et al.\cite{ xing2018intelligent} achieved impressive Sharpe ratios and minimized drawdown (MDD) by incorporating sentiment analysis, utilizing the Error-Correcting Mechanism (ECM) as a screening mechanism for LSTM models. Initially, the prediction of stock market comments began with the Bag-of-Words approach. This approach models news articles using a vector space model, which translates each news piece into a vector representing word statistical measures, such as word frequency. While several bag-of-words-based approaches have shown predictive power in previous studies\cite{ wu2008integrating}, they overlook a crucial element in the mapping from textual news articles to final directional predictions: sentiment analysis. Li et al. \cite{li2014news} conducted an analysis focused on the Hong Kong stock market using financial news English comments from 2003 to 2008, sourced from the FINET dataset. They manually created the Harvard IV-4 sentiment dictionary (HVD) and used the Loughran-McDonald Financial Sentiment Dictionary (LMD) for sentiment analysis. A word frequency matrix was constructed from the news comments and projected onto the sentiment space to create sentiment vectors using these constructed sentiment dictionaries. Daily stock opening and closing prices were used to calculate price volatility, which served as labels. Three label categories, positive, neutral, and negative, were generated based on various thresholds. The author employed a support vector machine (SVM) for cross-training and validation, concluding the release of financial news does indeed impact stock returns, though without quantifying this impact more effectively. Furthermore, the dataset consisted solely of English comments, rather than the mainstream traditional Chinese used in Hong Kong. Therefore, there is considerable room for improvement in these aspects. 

Nguyen et al. \cite{nguyen2015sentiment} introduced a method for incorporating social media sentiment into the prediction of stock price movements. This approach encompasses two distinct ways of capturing topic sentiment associations. The first method is based on JST (Joint Sentiment-Topic), which relies on pre-existing topic models. The second method utilizes an aspect-based sentiment approach, wherein the identification of topics and sentiments is facilitated by the proposed methodology. From a practical standpoint, this method demonstrates a noteworthy ability to forecast the stock price trends of select stocks, achieving an accuracy rate exceeding 60\%. Its performance surpasses that of alternative methods, especially for stocks that prove challenging to predict. However, the overall average accuracy of this method remains at 54.41\%. In addition, it should be noted that the implementation of aspect-based sentiment analysis in languages other than English may encounter stability and efficiency challenges.

Derakhshan \& Beigy \cite{derakhshan2019sentiment} introduced the LDA-POS method, which leverages part-of-speech information in the LDA model to separate words based on their part-of-speech tags. This model demonstrated notable results on two datasets, one in English and the other in Persian. The study utilized a comment dataset from Yahoo Finance covering the period from 2012 to 2013 and data from the Sahamyab website in Iran spanning from 2016 to 2018. The LDA-POS method combines the LDA topic model with part-of-speech tagging for comment analysis. It was compared to the Bag-of-Words (BOW) model and the sentiment polarity model. Experimental results demonstrated that, on both the validation and independent test sets, the LDA-POS method achieved higher average prediction accuracy compared to the BOW model. The prediction accuracy of the LDA-POS method on English and Persian datasets was comparable, suggesting its suitability for different languages. While sentiment analysis contributed to improved prediction accuracy, the LDA-POS method outperformed both the sole use of sentiment polarity and the BOW method. However, the obtained accuracy was not significantly high, with values of 56.24\% and 55.33\%, respectively \cite{derakhshan2019sentiment}. This could be attributed to the relatively small size of the dataset, which may have limited the model's ability to capture sufficient patterns and generalize effectively. 

\subsection{Sentiment analysis to predict stock trend}
The prediction of stock market trends has been a pivotal aspect of the finance industry since its inception. The standard methods utilized for prediction encompass technical analysis and fundamental analysis. Our study specifically concentrates on technical analysis, utilizing stock prices and various derived indicators, with a particular focus on sentiment analysis, to forecast market trends.

According to the study of Bustos and Pomares-Quimbaya \cite{bustos2020stock}, stock prices are influenced by an array of factors, encompassing economic background, sentiment indicators, search engine trend analysis, and a company's news. Financial media frequently correlates news with economic context, offering invaluable insights for predicting stock prices. While professional media provides context related to the economic environment, general news may lack explicit context yet still significantly impact stock prices. Social network and search engine analyses have demonstrated remarkable efficacy in predicting stock market trends. Studies endeavor to estimate public sentiment regarding stocks, showcasing the practicality of these indicators in predicting stock market trends.

Medhat et al. \cite{medhat2014sentiment} proposed the techniques for sentiment classification can be categorized into machine learning methods, lexicon-based methods, and hybrid methods. Machine learning methods leverage well-known algorithms and linguistic features. Lexicon-based methods rely on sentiment lexicons containing precompiled sentiment terms. These methods are further divided into dictionary-based and corpus-based approaches. Dictionary-based methods dedicate identifying opinion-seed words and exploring synonyms and antonyms in the dictionary. Conversely, corpus-based methods commence with a seed list of opinion words and identify additional opinion words in a substantial corpus to discern words with specific contextual orientations. These methods employ statistical or semantic techniques for analysis.

Li Q, et al.\cite{li2014effect} proposed a method to refine lexicons by analyzing the contextual information of financial domain vocabulary from the web. Researchers utilized optimistic and pessimistic public sentiments as inputs to their model, gauging their predictive power on stock market trends. The study revealed that pessimistic public sentiment significantly influenced stock market trends, whereas optimistic sentiment had a limited ability to perceive these trends. Furthermore, the research indicated that incorporating public sentiment into the prediction model within three to six days further enhanced the accuracy of stock predictions. Li, Wu et al. \cite{li2020incorporating} transformed stock prices into technical indicators representing historical stock prices. They analyzed sentiment categories in various financial news reports and constructed a two-layer LSTM network to grasp indicators and sequential information from news sentiment. The study utilized stock data from the Hong Kong Stock Exchange between January 2003 and March 2008 and corresponding FINET news from the same period as sentiment data. Experiments involved utilizing stock prices and news sentiment as inputs, employing diverse models and sentiment lexicons for stock market prediction. Researchers employed three models: LSTM, MKL, and SVM, along with four sentiment lexicons (SenticNet 5, SentiWordNet 3.0, Vader, and Loughran-McDonald Financial Dictionary 2018) to model news sentiment. These lexicons encompass sentiment polarity values and four sentiment dimensions: \emph{Pleasantness}, \emph{Attention}, \emph{Sensitivity}, and \emph{Aptitude}. Sentiment polarity values are indicators that reveal whether a word conveys a positive, neutral, or negative sentiment. The results demonstrate that this method outperforms MKL and SVM in terms of both prediction accuracy and F1 score. Moreover, models that integrate stock prices and news sentiments exhibit superior performance compared to models that solely rely on technical indicators or news sentiments, both at the individual stock level and sector level.

\subsection{Development of text analysis technology based on topic model}
The research on topic models can be traced back to the proposal of Indexing by Latent Semantic Analysis (LSA) by Deerwester et al. in 1990 \cite{deerwester1990indexing}. LSA operates by creating a "vector semantic space" to extract "concepts" from documents and words, analyzing their relationships. Its underlying assumption is that two words are semantically similar if they frequently appear in the same document. LSA builds an Occurrence Matrix using a substantial amount of text data, where each row represents a word, each column represents a document, and the matrix elements indicate the frequency of word occurrence in each document. To reduce dimensionality while preserving information, Singular Value Decomposition (SVD) is applied to the matrix. The similarity between two words can then be measured using the cosine value of their row vectors or the dot product of their normalized vectors. A value close to one indicates high similarity, while a value close to zero suggests low similarity. However, LSA has limitations, the most prominent being its weak statistical foundation. 

To address these limitations, Hofmann introduced an improved method called Probabilistic Latent Semantic Analysis (PLSA) in 1999 \cite{hofmann1999probabilistic}. PLSA models the word generation process in a document as a mixture of multiple implicit topics, each associated with a set of words. It employs maximum likelihood estimation for parameter estimation and provides a stronger statistical foundation compared to LSA. Additionally, PLSA excels in handling polysemous words by establishing a latent semantic space and demonstrates better results in dealing with domain-specific synonyms. 

Nonetheless, the PLSA model has certain limitations in terms of model inference and the selection of topic distributions since it does not incorporate prior distributions. To overcome these limitations, David et al. proposed Latent Dirichlet Allocation (LDA) in 2003 \cite{blei2003latent}, which extends the PLSA model by introducing a Dirichlet prior to the document-topic and topic-word distribution. Both LDA and PLSA are generative probabilistic models that discretize the continuous topic space into k topics and represent documents as combinations of these k topics. The topic distributions in these models are determined by calculating the probability of each word appearing in each topic and the probability of each document containing each topic. Early conventional topic models, such as LSA, PLSA, and LDA, rely on the bag-of-words representation of text, disregarding word order and deeper meanings, which limits their representational capabilities. Furthermore, in PLSA and LDA, the number of topics (K) must be assumed to be known during the modeling process, and an incorrect choice of K can affect the accuracy of the results. Estimating the appropriate number of topics for large or unfamiliar datasets is challenging.

Moreover, both PLSA and LDA fail to differentiate between informative and less informative words, as they merely reproduce the distribution of words within documents. This lack of distinction results in outcomes that are evidently lacking in validity. Egger and Yu \cite{egger2021identifying} pointed out that although LDA explores a wide range of topics, it often exhibits a high repetition rate of content. On the other hand, Chen et al. \cite{chen2019experimental} highlighted the insufficient statistical learning property of the LDA model, making it less suitable for noisy and sparse datasets.

In addition to the bag-of-words model, the Non-Negative Matrix Factorization (NMF) model, introduced by Lee et al. in 1999, is another widely utilized topic model \cite{lee1999learning}. NMF is an approach to topic modeling based on matrix decomposition. It assumes that both the document matrix and topic matrix are non-negative and extracts topics by decomposing the document matrix into two non-negative matrices: the document-topic matrix and the topic-word matrix. NMF is effective in extracting concise textual data. However, it is prone to generating broad topics that lack specificity. 

In recent years, the integration of deep learning with topic models has emerged as a promising direction, offering potential solutions to address the limitations of traditional topic modeling approaches. For instance, in 2020, Angelov proposed the TopicVec model \cite{angelov2020top2vec}, which serves as a neural network-based distributed topic model, effectively addressing the aforementioned challenges. TopicVec leverages Doc2Vec technology to map documents and words into a shared semantic vector space. The dimensionality of the document vectors is then reduced using UMAP, thereby enhancing the effectiveness and efficiency of subsequent clustering algorithms and avoiding the curse of dimensionality. HDBSCAN is employed to cluster the document vectors, leading to distinct clusters that represent different topics. The topic representation is obtained by averaging the vectors from documents within the same cluster, and the top N words closest to the topic vector are selected as the topic representation. Notably, TopicVec automatically determines the number of topics through HDBSCAN, providing a more informative representation of topics compared to the traditional bag-of-words approach.

However, it is important to acknowledge that Top2Vec has certain limitations. It utilizes the density-based clustering method HDBSCAN for clustering, but when determining the topic vectors, it relies on a calculation method based on the centroid of each dense region of the document vector. This approach may result in less accurate topic vectors and representations.

To overcome the limitations of Top2Vec, the BERTopic model \cite{grootendorst2022bertopic} adopts a distinct approach by separately embedding documents and words using pre-trained models, eliminating the need to specify the number of topics in advance. The clustering process is then optimized by reducing the dimensionality of the embeddings before performing clustering on these reduced-dimensional embeddings. Once different topics are obtained through dimensionality reduction and clustering, all documents belonging to the same topic are merged into one comprehensive document. Within this merged document, the words with the highest c-TF-IDF values are selected as the topic representations. This approach essentially employs a bag-of-words strategy to identify topic representations by considering the frequency of each word in the respective topic category and multiplying it with the inverse document frequency of the word in the entire corpus to determine its weight. It is worth mentioning that BERTopic shares a similar methodology with Top2Vec in terms of dimensionality reduction and clustering, which will not be reiterated here.

The innovation and advantages of BERTopic are evident. Unlike traditional topic models, BERTopic can automatically determine the number of topics and supports hierarchical topic reduction. Compared to Top2Vec, the use of pre-trained models enhances its representation capabilities, while the separation of document embedding encoding and topic representation provides greater flexibility.

\subsection{Application status and effect evaluation of BERTopic}
BERTopic \cite{grootendorst2022bertopic}, an algorithm based on the pre-trained language model BERT, has been employed to computationally analyze a vast amount of subjective information data.BERTopic has demonstrated its robust utility in the realm of social media data analysis, furnishing valuable insights into user preferences, social connections, and emerging subject matter. It has emerged as a prominent method for automating topic extraction across a spectrum of applications, encompassing sentiment analysis and recommender systems. This capacity allows for precise topic modeling even when confronted with subpar text quality. Grootendorst \cite{grootendorst2022bertopic} substantiated the efficacy of BERTopic through experiments conducted on three datasets: 20 Newsgroups, BBC News, and President Trump’s tweets. 

The research evaluated BERTopic using a thematic consistency metric, which unveiled consistent themes spanning the diverse datasets, thereby demonstrating its outstanding performance. In a separate study, Ogunleye et al. \cite{app13020797} employed a dataset comprising tweets from Nigerian bank customers and introduced KernelPCA and K-means clustering into the BERTopic framework. The findings indicated that this approach yielded coherent themes. Beyond its role in social media data analysis, BERTopic has also found application in the retrieval of document topics. Silveira et al. \cite{silveira2021topic} developed a topic model for legal documents by employing a stochastic topic modeling approach that successfully integrated BERTopic for aspect extraction. 

BERTopic demonstrates superior performance compared to earlier topic models when dealing with social media data. Jaradat et al. \cite{jaradat2019dynamic} analyzed Twitter data and found that LDA models often overlook co-occurrence relationships, even when multiple topics exist in a document simultaneously. On the other hand, BERTopic, an emerging topic modeling technique, leverages vector space to approximate similar text, resulting in improved interpretability and resource efficiency.

In a study conducted by Abuzayed and Al-Khalifa \cite{abuzayed2021bert}, experiments were conducted using Arabic documents from three Arabic online newspapers (Assabah, Hespress, and Akhbarona), and the results were compared with those from LDA and NMF. The findings indicated that BERTopic outperforms LDA and NMF, showcasing its adaptability to different topic sizes. Similarly, Raju et al. \cite{raju2022topic} evaluated the effectiveness of LDA, LSA, and BERTopic in topic modeling using a dataset extracted from the Consumer Financial Protection Bureau. They consistently found that BERTopic achieved excellent consistency scores.

In another study, Hutama et al. \cite{hutama2022indonesian} utilized pre-trained multilingual models (XLM-R and mBERT) combined with the BERTopic model to classify Indonesian hoax news. The proposed method outperformed the baseline model in classifying hoax news in Indonesian, a low-resource language, with accuracy, precision, recall, and F1 results of 0.9051, 0.9515, 0.8233, and 0.8828, respectively.

Furthermore, Hristova et al. \cite{hristova2022media} employed the BERTopic model for dynamic topic modeling using data samples from 1200 different Bulgarian news websites and over 71,000 news comments crawled from Bulgaria's popular community platform, "www.bg-mamma.com." Their experiments demonstrated that BERTopic effectively reduces computation time and noise in crawled text data.

With the aim of extracting themes from online platforms frequented by stockholders, BERTopic can automatically extract themes from text data and cluster texts with similarthemes. Given the significant advancements in identifying fake news, it is conceivable that BERTopic may assist in analyzing 
sentiment in comments and lead to improved predictions for stock market fluctuations.

\section{Methodology}

\subsection{Data Collection}
The data set contains tweets for top 25 most watched stock tickers on Yahoo Finance from 30-09-2021 to 30-09-2022, additionally was added stock market price and volume data for corresponding dates and stocks.\\
	\begin{itemize}
		\item \textbf{Date} - date and time of tweet
		\item \textbf{Tweet} - full text of the tweet
		\item \textbf{Stock Name} - full stock ticker name for which the tweet was scraped
		\item \textbf{Company Name} - full company name for corresponding tweet and stock ticker
	\end{itemize}

\begin{table*}
	\renewcommand{\arraystretch}{2}
	\setlength{\tabcolsep}{3pt} 
	\scriptsize 
	\caption{An Example of Comments for Amazon (AMZN)}
	\label{tab:comment}
	\begin{tabularx}{\textwidth}{|c|c|X|c|c|}
		\hline
		\textbf{Index} & \textbf{Date} & \textbf{Tweet} & \textbf{Stock Name} & \textbf{Company Name} \\ 
		\hline
		48351 & 2022-09-29 22:40:47+00:00 & A group of lawmakers led by Sen. Elizabeth War... & AMZN & Amazon.com, Inc. \\ 
		\hline
		48352 & 2022-09-29 22:23:54+00:00 & \$NIO just because I'm down money doesn't mean ... & AMZN & Amazon.com, Inc. \\ 
		\hline
		48353 & 2022-09-29 18:34:51+00:00 & Today’s drop in \$SPX is a perfect example of w... & AMZN & Amazon.com, Inc. \\ 
		\hline
		48354 & 2022-09-29 15:57:59+00:00 & Druckenmiller owned \$CVNA this year \newline Munger b... & AMZN & Amazon.com, Inc. \\ 
		\hline
		48355 & 2022-09-29 15:10:30+00:00 & Top 10 \$QQQ Holdings \newline \newline And Credit Rating & AMZN & Amazon.com, Inc. \\ 
		\hline
	\end{tabularx}
\end{table*}
\begin{table}[H]
	\centering
	\caption{An Example of Stock Data for Amazon (AMZN)}
	\label{tab:stock_data}
	\resizebox{0.5\textwidth}{!}{%
		\begin{tabular}{|c|c|c|c|c|c|c|c|}
			\hline
			\textbf{Date} & \textbf{Open} & \textbf{High} & \textbf{Low} & \textbf{Close} & \textbf{Adj Close} & \textbf{Volume} & \textbf{Stock Name} \\ 
			\hline
			2021-09-30 & 165.80 & 166.39 & 163.70 & 164.25 & 164.25 & 56848000 & AMZN \\ 
			\hline
			2021-10-01 & 164.45 & 165.46 & 162.80 & 164.16 & 164.16 & 56712000 & AMZN \\ 
			\hline
			2021-10-04 & 163.97 & 164.00 & 158.81 & 159.49 & 159.49 & 90462000 & AMZN \\ 
			\hline
			2021-10-05 & 160.23 & 163.04 & 160.12 & 161.05 & 161.05 & 65384000 & AMZN \\ 
			\hline
			2021-10-06 & 160.68 & 163.22 & 159.93 & 163.10 & 163.10 & 50660000 & AMZN \\ 
			\hline
			2021-10-07 & 164.58 & 166.29 & 164.15 & 165.12 & 165.12 & 48182000 & AMZN \\ 
			\hline
			2021-10-08 & 165.85 & 166.07 & 164.41 & 164.43 & 164.43 & 39964000 & AMZN \\ 
			\hline
			2021-10-11 & 163.75 & 164.63 & 161.90 & 162.32 & 162.32 & 40684000 & AMZN \\ 
			\hline
			2021-10-12 & 162.85 & 163.38 & 161.81 & 162.37 & 162.37 & 36392000 & AMZN \\ 
			\hline
		\end{tabular}%
	}
\end{table}

Additionally, the stock price market price data is shown in Table \ref{tab:stock_data}. Following the data processing phase, sentiment analysis was conducted on the comments to extract their sentiment values, and topic modeling was employed to identify the topics they pertain to. Subsequently, a dataset was constructed incorporating these attributes. The descriptions of these attributes are provided below:

\begin{itemize}
	\item \textbf{Date}\\
	The specific day on which the stock data was recorded.
	
	\item \textbf{Open}\\
	The price of the stock when the market opened on the given date.
	
	\item \textbf{High}\\
	The highest price at which the stock traded during the course of the day.
	
	\item \textbf{Low}\\
	The lowest price at which the stock traded during the day.
	
	\item \textbf{Close}\\
	The price of the stock when the market closed on the given date.
	
	\item \textbf{Adj Close}\\
	The closing price adjusted for factors such as dividends, stock splits, and new stock offerings, providing a more accurate reflection of the stock's value.
	
	\item \textbf{Volume}\\
	The number of shares of the stock that were traded during the day.
	
	\item \textbf{MA7}\\
	The 7-day moving average of the stock's closing price. It's the average of the closing prices over the last 7 days.
	
	\item \textbf{MA20}\\
	The 20-day moving average of the stock's closing price. It's the average of the closing prices over the last 20 days.
	
	\item \textbf{MACD} (Moving Average Convergence Divergence)\\
	A trend-following momentum indicator that shows the relationship between two moving averages of a security’s price, typically the 12-day and 26-day exponential moving averages.
	
	\item \textbf{20SD}\\
	The standard deviation of the stock's closing price over the last 20 days, indicating the price's volatility.
	
	\item \textbf{upper\_band}\\
	The upper Bollinger Band, calculated as the sum of the 20-day moving average (MA20) and the product of 20SD and a chosen factor (commonly 2). It represents a higher range of stock price volatility.
	
	\item \textbf{lower\_band}\\
	The lower Bollinger Band, calculated as the difference between the 20-day moving average (MA20) and the product of 20SD and a chosen factor (commonly 2). It represents a lower range of stock price volatility.
	
	\item \textbf{EMA (Exponential Moving Average)}\\
	A type of moving average that places more weight on the most recent data points, making it more responsive to price changes.
	
	\item \textbf{log\_momentum}\\
	The natural logarithm of the momentum of the stock, where momentum is typically calculated as the difference between the current closing price and the closing price of a specified number of days ago.
	
	\item \textbf{score}\\
	The mean sentiment valuation derived from the aggregation of comments pertaining to a specific stock on a given day.
	
	\item \textbf{score\_topic}\\
	 The mean sentiment valuation derived from the sentiment scores associated with the topics of each comment for a specific stock on a given day.
\end{itemize}

\subsection{Sentiment Model}
We will conduct experiments employing both the VADER and BERT methodologies to assign scores to the texts. Subsequently, a comparative analysis of the performance of these two methods will be undertaken.
\subsubsection{Valence Aware Dictionary and Sentiment Reasoner (VADER)}
VADER is a lexicon and rule-based sentiment analysis tool specifically designed to discern sentiment from text, especially for social media and short-text content. Unlike traditional sentiment analysis models that often require extensive training data, VADER is pre-trained and can be applied directly to raw text, making it particularly useful for analyzing vast amounts of unstructured data.

Key Features:
\begin{itemize}
\item \textbf{Lexicon}: VADER's lexicon contains over 7,500 features, including words, emoticons, and slang, each annotated with its sentiment valence. The valence score ranges from -4 (most negative) to 4 (most positive).
\item \textbf{Incorporation of Grammatical and Syntactical Rules}: VADER doesn't solely rely on the lexicon. It also considers grammatical structures, intensifiers, and the context of words. For instance, the presence of an exclamation mark can amplify the sentiment intensity.
\item \textbf{Handling of Negations}: VADER is adept at understanding negations, which can reverse the sentiment of a statement. For example, "not good" is recognized as negative despite the presence of the positive word "good".
\item \textbf{Emphasis on Capitalization}: The tool can differentiate between words based on their capitalization. For instance, "GOOD" might be perceived as more positive than "good".
\end{itemize}

\subsection{BERTopic}

\subsubsection{Sentence Embeddings}
Machine Learning algorithms are not inherently designed to handle unprocessed textual information. As a result, a crucial step involves converting these texts into numerical representations, commonly referred to as embeddings. Predominantly, we employ Sentence-BERT (SBERT) for this purpose, generating embeddings for each document. It's worth highlighting the distinction between BERT and SBERT: while the former produces word-level embeddings, the latter yields a consolidated embedding vector for an entire sentence. A significant advantage of SBERT is its specialized training for semantic similarity. Furthermore, its seamless integration with the SentenceTransformers library ensures its adaptability with models of a similar nature.

\subsubsection{Dimensionality Reduction}
Subsequent to obtaining embeddings, the next step involves clustering them. However, prior to this, a reduction in the dimensionality of the embedding vectors is essential. The rationale behind this is the susceptibility of many clustering algorithms to the 'curse of dimensionality'. In essence, when data points are dispersed across numerous dimensions, they tend to become equidistant from one another, complicating the clustering process. To address this, we employ the Uniform Manifold Approximation and Projection (UMAP) technique, a dimensionality reduction method akin to PCA (Principal Component Analysis) and t-SNE.\\
The first term acts as an ``attractive force" whenever there is a large weight associated to the high-dimensional case. This is because this term will be minimized when $w_l(e)$ is as large as possible, which will occur when the distance between the points is as small as possible. The second term acts as a ``repulsive force" whenever high-dimensional weight $w_h(e)$ is small. This is because the term will be minimized by making $w_l(e)$ as small as possible.
\begin{equation}
	CE=\sum_{e\in E}{w_{h}(e)}\log(\frac{w_{h}(e)}{w_{l}(e)})+(1-w_{h}(e))\log(\frac{1-w_{h}(e)}{1-w_{l}(e)})
\end{equation}
\subsubsection{Clustering}
Upon dimensionality reduction, the clustering algorithm HDBSCAN (Hierarchical DBSCAN) is applied to the condensed embedding vectors. This is the default choice, though other algorithms are also explored in our research. Contrary to k-means, which is distance-centric, HDBSCAN operates based on density, forming clusters contingent on point reachability. Points that are significantly distant from the majority are designated as outliers. Our study also incorporates a variety of other clustering algorithms for comparative analysis.
HDBSCAN (Hierarchical Density-Based Spatial Clustering of Applications with Noise) is an advanced clustering algorithm that extends the DBSCAN algorithm to work with varying densities. 

Steps of HDBSCAN:
\begin{enumerate}
	\item Transform the space
		\begin{itemize}
			\item Convert the original space of data points into a space that reflects the density/sparsity of the data.
			\item This is done by computing the "mutual reachability distance" between each pair of data points.
 		\end{itemize}
		\item Build the Minimum Spanning Tree (MST)\\
		Construct a MST of the data points using the mutual reachability distance as the edge weight between any pair of points.
		
		\item Construct a hierarchy of clusters\\ 
		This is done by iteratively connecting the closest pair of clusters (in terms of edge weight) and treating them as a single cluster in subsequent iterations.
		
		\item Condense the cluster hierarchy\\
		Create a condensed tree from the cluster hierarchy, which only retains the most stable clusters. Stability of a cluster is determined by its persistence over the hierarchy.
		
		\item Extract clusters\\
		Extract the clusters from the condensed tree based on a stability threshold.
		\end{enumerate}

Principle by Formula:
\begin{enumerate}

\item Mutual Reachability Distance
Given two points \(p\) and \(q\), the mutual reachability distance is defined as:
\begin{equation}
	\small d_{mreach}(p,q)=\max\!\{\mathrm{core\text{-}dist}(p),\,\mathrm{core\text{-}dist}(q),\,d(p,q)\}
\end{equation}
Where:
\begin{itemize}
	\item  $d(p, q)$ is the original distance between points $p$ and $q$.
	\item $core-dist(p)$ is the distance from point $p$ to its $MinPts^{th}$ nearest neighbor.
\end{itemize}
\item Core Distance
For a point $p$, given a minimum number of points $MinPts$, the core distance is the distance from $p$ to its \(MinPts^{th}\) nearest neighbor. It is defined as:
\begin{equation}
	core-dist(p) = d(p, o)
\end{equation}
Where:
\begin{itemize}
	\item $o$ is the $MinPts^{th}$ nearest neighbor of $p$.
 \end{itemize}
	\item Stability
Stability of a cluster is a measure of its persistence over the hierarchy. It's calculated based on the birth and death levels in the hierarchy. The exact formula can be complex and depends on the specific implementation of HDBSCAN.

\end{enumerate}
These formulas and steps provide a high-level overview of HDBSCAN. The algorithm's actual implementation can be more intricate, especially when optimizing for computational efficiency.
\subsubsection{Topic Generation}
Post clustering, we are equipped with a collection of vectors and their corresponding cluster assignments. The objective now is to generate topics that encapsulate the most salient words, elucidating the essence of the topic. To realize this, we employ a variant of TF-IDF, termed c-TF-IDF. For this methodology, documents within a specific cluster are amalgamated through concatenation. Subsequently, the relative significance of each word is ascertained using its frequency within the current cluster, adjusted by the logarithm of its inverse cluster frequency. This inverse frequency metric is computed as the total number of clusters divided by the frequency of a word's appearance across all clusters. The inversion ensures that words ubiquitous across clusters have their importance attenuated, emphasizing words unique to specific clusters. Ultimately, words pivotal to a cluster or topic are identified based on their c-TF-IDF scores. These formulas computes the importance of a term $t$ within a specific class $c$. Terms that are frequent in a particular class but rare in other classes will have a high c-TF-IDF score, indicating their significance to that class.
\begin{enumerate}
	\item \textbf{Term Frequency (TF)} \\
	Given a term $t$ and a class $c$, the term frequency $TF(t, c)$ is defined as:
	\begin{equation}
		TF(t, c) = \text{Number of times term } t \text{ appears in class } c
	\end{equation}
	
	\item \textbf{Document Frequency (DF)} \\
	Given a term $t$, the document frequency $DF(t)$ is defined as:
	\begin{equation}
		DF(t) = \text{Number of classes containing term } t
	\end{equation}
	
	\item \textbf{Inverse Class Frequency (ICF)} \\
	Given a term $t$ and the total number of classes $N$, the inverse class frequency $ICF(t)$ is calculated as:
	\begin{equation}
		ICF(t) = \log\left(\frac{N}{1 + DF(t)}\right)
	\end{equation}
	
	\item \textbf{c-TF-IDF} \\
	Given a term $t$ and a class $c$, the c-TF-IDF is defined as:
	\begin{equation}
		c-TF-IDF(t, c) = TF(t, c) \times ICF(t)
	\end{equation}
\end{enumerate}

\begin{figure}[hp]
	\centering
	\includegraphics[width=90mm]{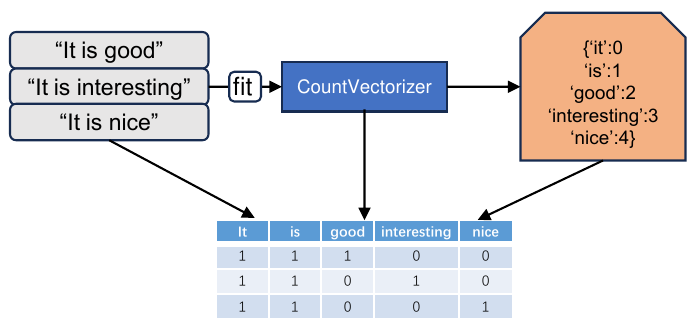}
	\caption{The usage of CountVectorizer}
	\label{countvectorizer}
\end{figure}

\begin{figure*}[htbp] 
	\centering
	\includegraphics[scale=0.5]{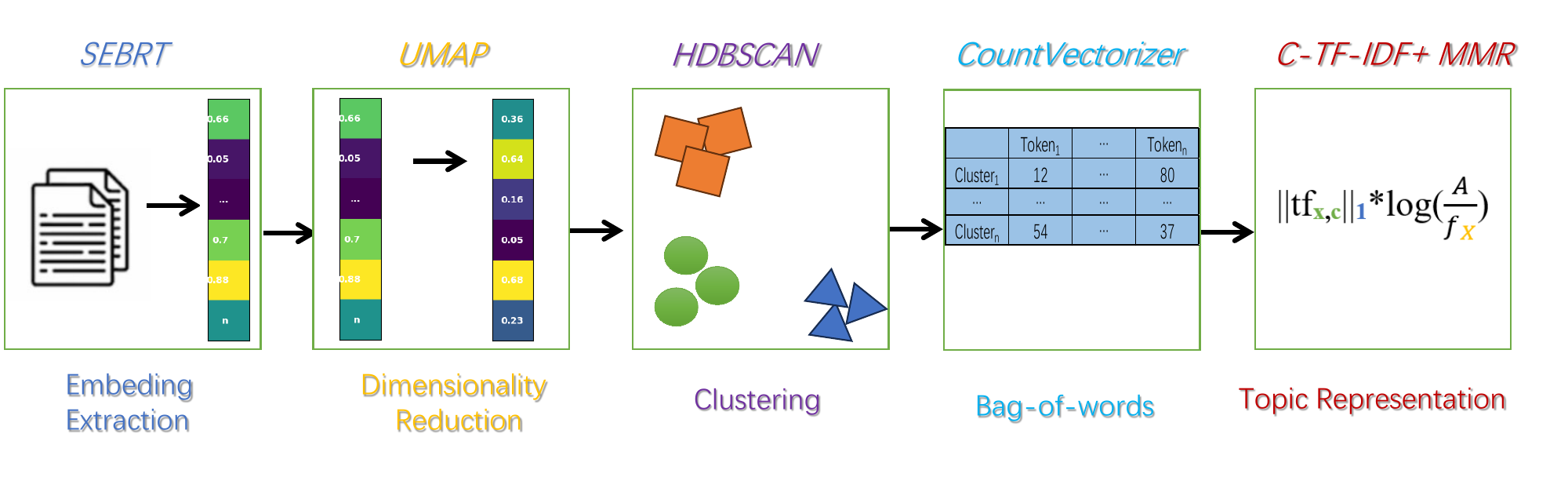}
	\caption{The structure of BERTopic}
	\label{bertopic}
\end{figure*}

The complete architecture of BERTopic is illustrated in Fig. \ref{bertopic}..

\subsection{LSTM}
\begin{figure}[H]
	\centering
	\includegraphics[scale=0.4]{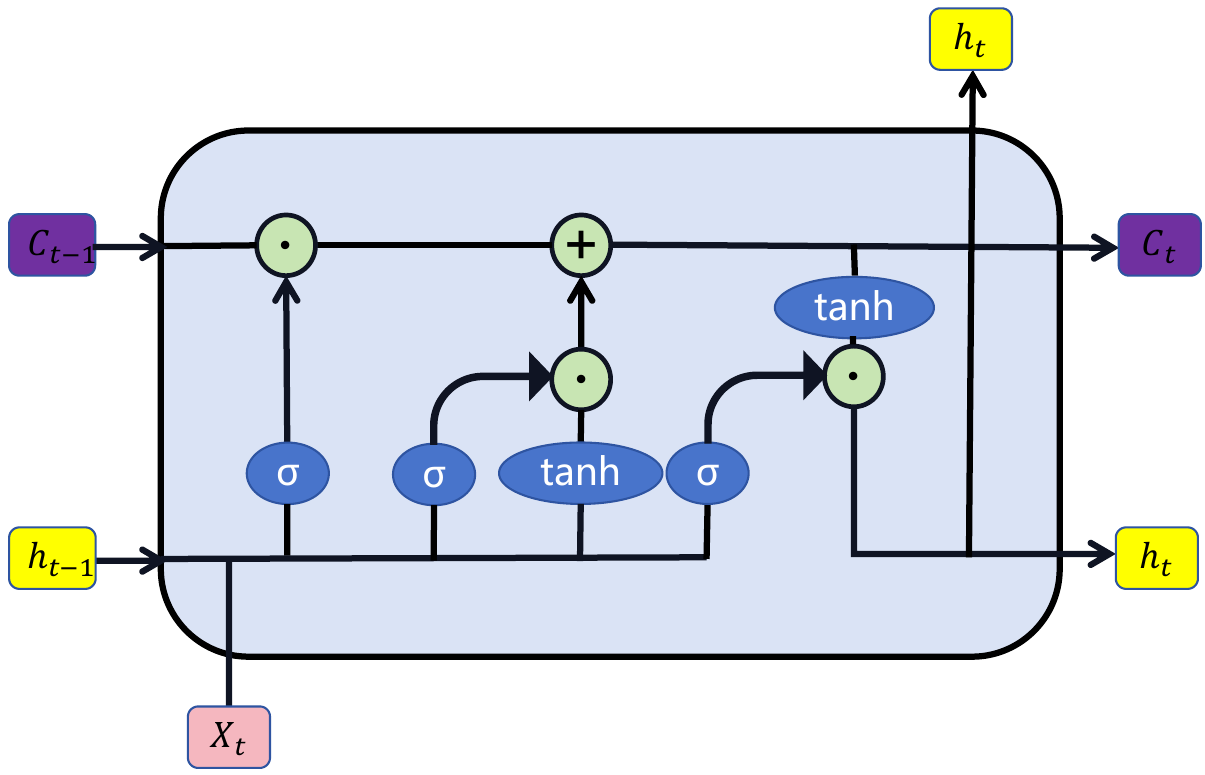}
	\caption{Long Short-Term Memory network.}
	\label{lstm}
\end{figure}

Long Short-Term Memory Networks (LSTM) are a specialized type of Recurrent Neural Network (RNN) that excel in processing sequential data. The LSTM architecture was initially introduced by Hochreiter and Schmidhuber in 1997. It addresses the issue of long-term dependency problems, which was a limitation of traditional RNNs. Consequently, LSTM has gained significant popularity and is widely employed in various domains including speech recognition, image description, and natural language processing. For reference, please see Fig. \ref{lstm}, which depicts the fundamental structure of LSTM.

A basic LSTM unit consists of a unit, a foget gate, an input gate and an output gate. The three gates regulate the flow of information related to the cell, protecting and controlling the cell state. The following is the structural process of LSTM.
We first give the meanings of some letters and symbols to introduce the structure of LSTM in more detail. $W_{f,x},W_{f,h},W_{\bar{c},x},W_{\bar{C},h},W_{i,x},W_{i,h},W_{o,x}$ and $W_{o,h}$ are weight matrices. $b_f$, $b_C$, $b_i$ and $b_o$ are bias vectors. $f_t$, $i_t$ and $o_t$ are vectors for the activation values of the respective gates. $\tilde{C_t}$ and $C_t$ are vectors for the cell states and candidate values. $h_t$ is a vector for the output of the LSTM layer. 

\subsubsection{The forget gate}
\begin{figure}[H]
	\centering
	\includegraphics[scale=0.4]{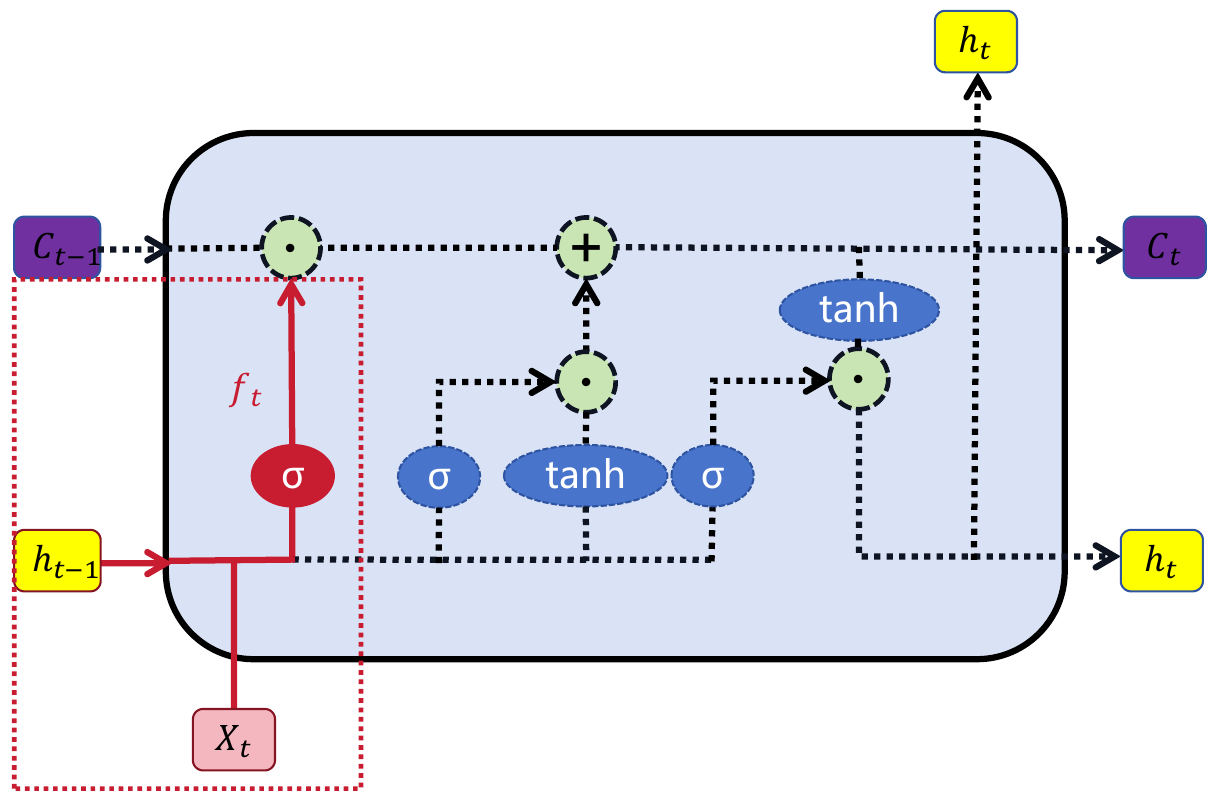}
	\caption{Forget Gate.}
	\label{forget}
\end{figure}

The forget gate defines what information is deleted from the cell state, that is, it determines what information is discarded from the cell state. This gate reads  $h_(t-1)$ sum $X_t$, and then after passing through the sigmoid ($\sigma$) layer, outputs an activation value between 0-1 $f_t$. Its is 0 means completely discarded, 1 means completely retained.
\begin{equation}
	f_{t}=\sigma(W_{f,x}x_{t}+W_{f,h}h_{t-1}+b_{f})
\end{equation}
The structure of the forget gate is shown in the Fig. \ref{forget}.

\subsubsection{The input gate}
The input gate determines what information should be added to the network's cell state ($C_t$). The process consists of two operations: First, candidate values $\tilde{C}_t$ are calculated that might be added to the cell state. Secondly, calculate the activation value it of the input gate. The architecture is showed in Fig. {input}. The formula structure of the input gate are as follows:
\begin{equation}
	\widetilde C_{t}=tanh(W_{\widetilde C,x}x_{t}+W_{\widetilde C,h}h_{t-1}+b_{\widetilde C})
\end{equation}
\begin{equation}
	i_t=\sigma(W_{i,x}x_t+W_{i,h}h_{t-1}+b_i)
\end{equation}
\begin{figure}[H]
	\centering
	\includegraphics[scale=0.4]{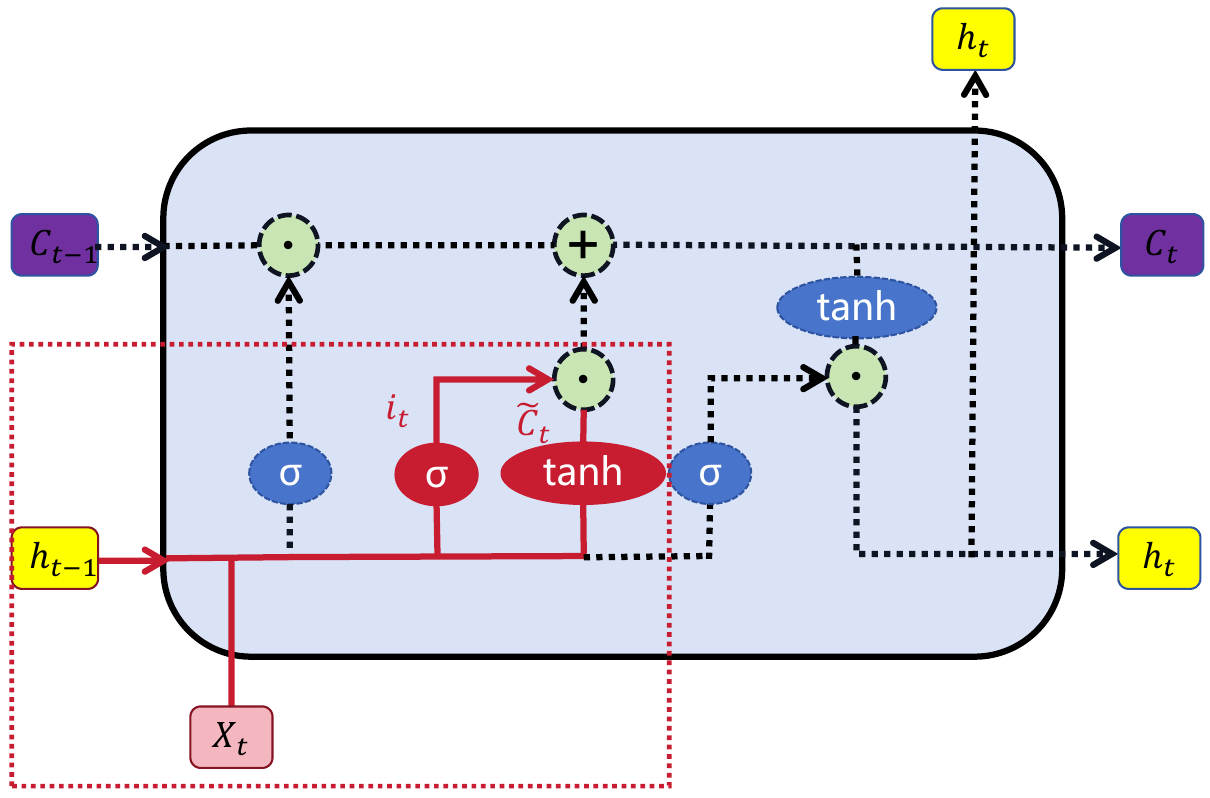}
	\caption{Input Gate.}
	\label{input}
\end{figure}

\subsubsection{Update the cell status.}
\begin{figure}[H]
	\centering
	\includegraphics[scale=0.4]{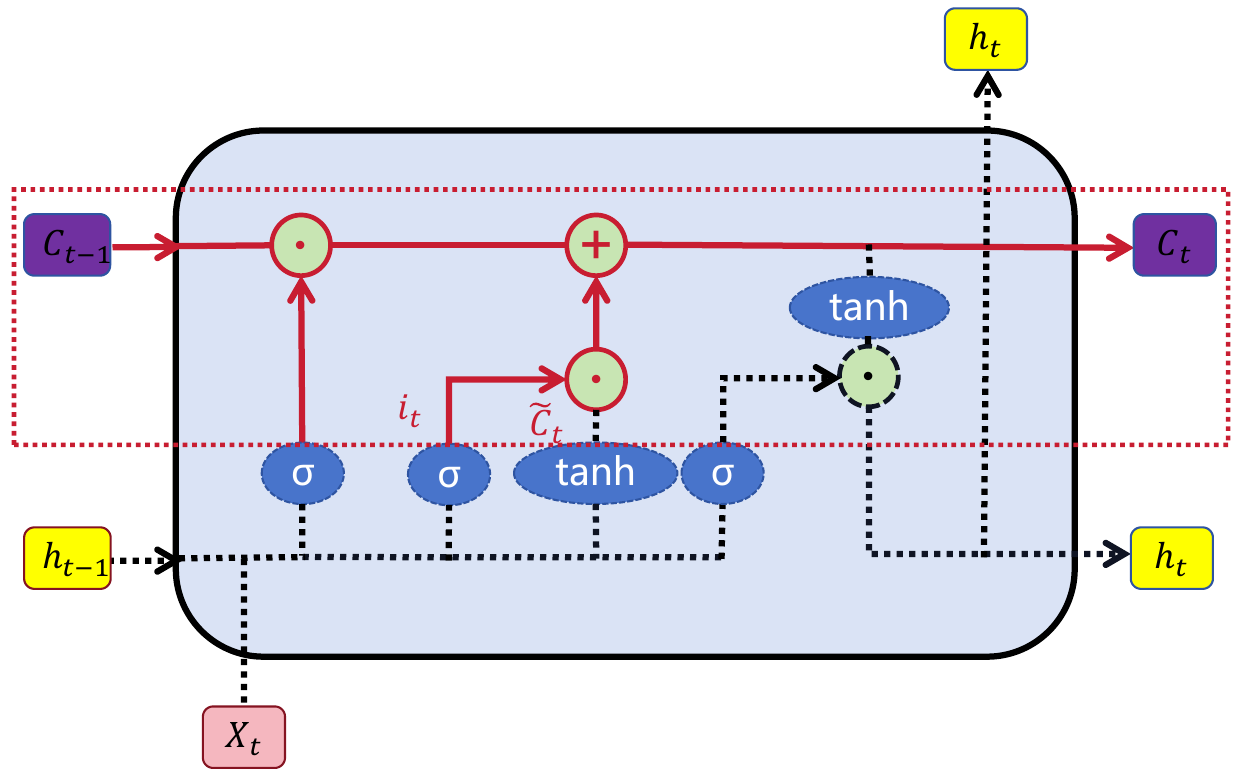}
	\caption{Update the cell status.}
	\label{update}
\end{figure}

As is showed in Fig \ref{update}. Where $\odot$ represents Handmard Product.
\begin{equation}
	C_{t}=f_{t}\odot C_{t-1}+i_{t}\odot\tilde{C}_{t}
\end{equation}

\subsubsection{Output Gate}
\begin{figure}[H]
	\centering
	\includegraphics[scale=0.4]{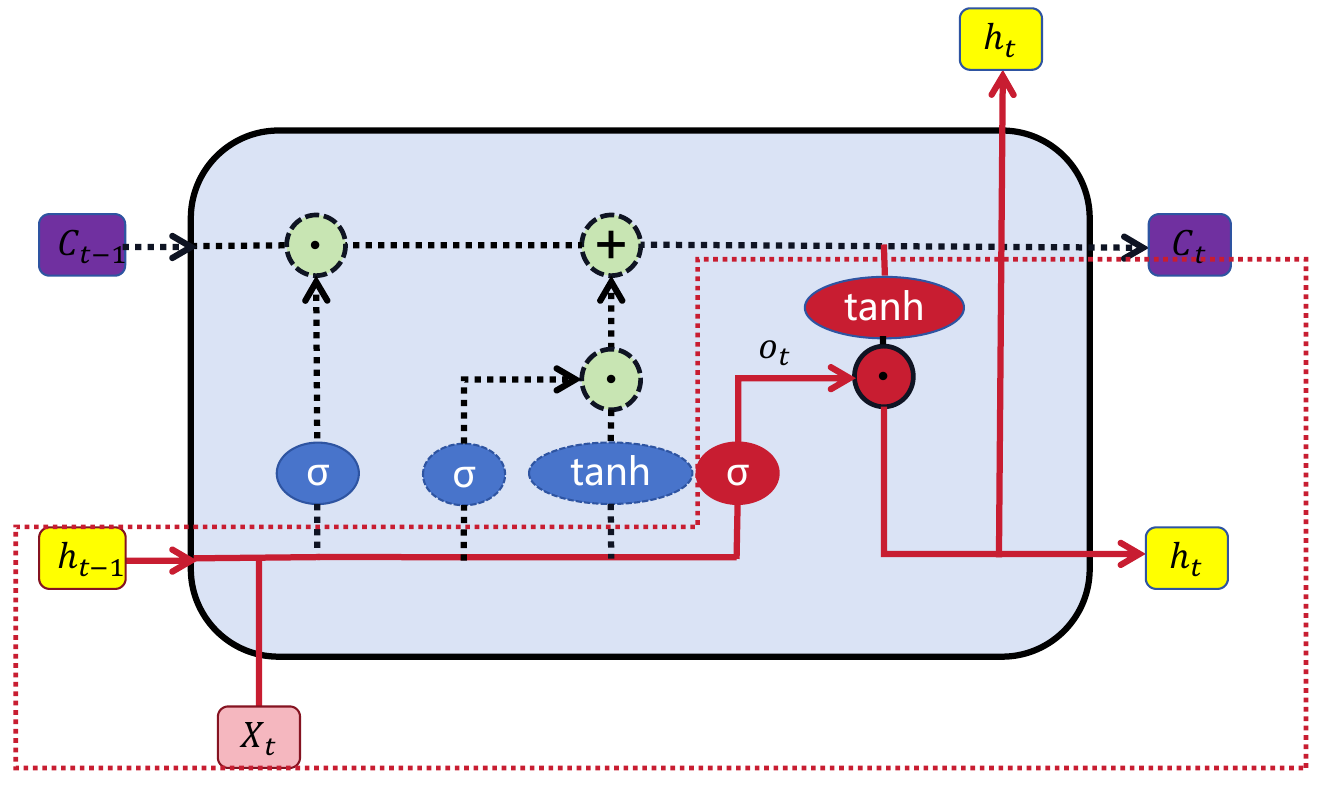}
	\caption{The output gate.}
	\label{output}
\end{figure}

Finally, the output gate $o_t$ and the hidden state $h_t$ are calculated. First, input $X_t$ and the last hidden state $h_(t-1)$ to calculate the value of the output gate through the sigmoid layer. Then, the cell state $C_t$ performs a tanh operation and is multiplied with the output gate to obtain the hidden layer state of the unit. As is showed in Fig. \ref{output}. Their formulas and structures are as follows:
\begin{equation}
	C_{t}=f_{t}\odot C_{t-1}+i_{t}\odot\tilde{C}_{t}
\end{equation}
\begin{equation}
	h_{t}=o_{t}\odot tanh(C_{t})
\end{equation}

\subsection{CNN}
\begin{figure}[H]
	\centering
	\includegraphics[scale=0.25]{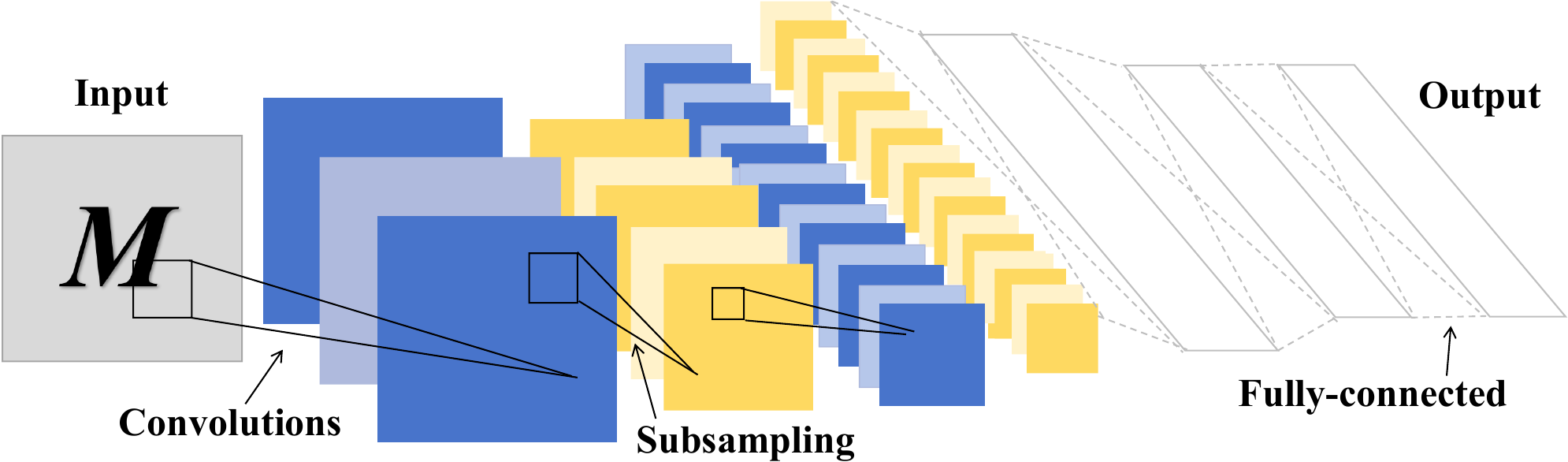}
	\caption{The LeNet-5 network architecture, which demonstrates excellent performance in classifying digits.}
	\label{cnn}
\end{figure}

A convolutional neural network (CNN) is composed of three main types of layers: convolutional layers, pooling layers, and fully connected layers. These layers are arranged in a stacked manner to form the architecture of a CNN as is showed in Fig. \ref{cnn}.

The convolutional layer plays a key role in determining the output of neurons connected to localized regions of the input. It achieves this by calculating the scalar product between the weights of the neurons and the corresponding localized regions of the input.
The pooling layer, on the other hand, performs subsampling on the spatial dimensions of a given input. This operation helps to reduce the number of parameters in the activation, resulting in a more compact representation of the features.
Following the convolutional and pooling layers, the feature maps can be flattened into a one-dimensional vector, which is then connected to the fully connected layer. In the fully connected layer, the network learns the weights and biases to establish the mapping relationship between the input and output.
By employing convolution and subsampling techniques, the CNN is capable of transforming the raw input data layer by layer. This transformation process enables the network to generate category scores for classification and regression tasks. Furthermore, this method allows the network to capture spatial localization and feature hierarchies in the input data, facilitating the extraction of meaningful features and making predictions that are relevant to the given task.

\subsubsection{Convolutional layer}
\begin{figure}[H]
	\centering
	\includegraphics[scale=0.3]{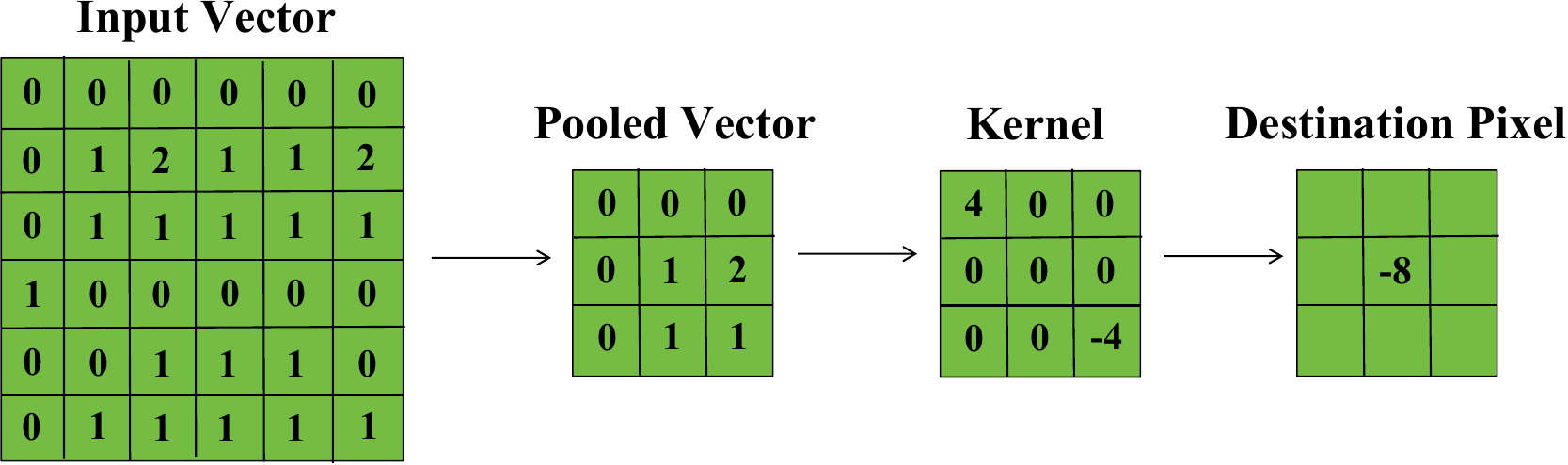}
	\caption{Convolutional layer.}
	\label{A visual representation of a convolutional layer. The center element of the kernel is placed on top of the input vector, which is then computed and replaced with a weighted sum of itself and nearby pixels.}
	\label{cnnlayer}
\end{figure}

As depicted in Fig. \ref{cnnlayer}, the convolutional layer of a neural network comprises multiple convolutional kernels, each responsible for computing a distinct feature map. In the feature map, every neuron is connected to a neighboring region in the previous layer known as the receptive field. To generate a new feature map, the input undergoes convolution with the corresponding learned convolution kernel, and the resulting output is obtained by applying a nonlinear activation function. In this process, all spatial locations of the input share the same convolution kernel, and by utilizing multiple different convolution kernels, a complete set of feature maps can be obtained.

Mathematically, the feature $z_(i,j,k)^l$ of the $k_{th}$ feature map position $(i,j)$ of layer $l$ is defined by the following equation:
\begin{equation}
	z_{i,j,k}^{l}={w_{k}^{l}}^{T}x_{i,j}^{l}+b_{k}^{l}
\end{equation}
where $w_k^l$ and $b_k^l$ are the weight vector and bias term of the $k_{th}$ filter in layer $l$ , respectively, and $x_(i,j)^l$ is the input patch at position $(i,j)$ in layer $l$. The activation function introduces non-linearity to the CNN and facilitates the detection of nonlinear features in multilayer networks. Let $\alpha$ denote the nonlinear activation function. The activation value of the convolutional feature $z_(i,j,k)^l$ can be computed as:
\begin{equation}
	\alpha_{i,j,k}^{l}=\alpha(z_{i,j,k}^{l})
\end{equation}

\subsubsection{Pooling layer}
After the convolution operation, the resulting feature map contains a multitude of distinct features. However, the abundance of features can potentially lead to overfitting issues. To address this concern, a commonly employed solution is to incorporate a pooling layer between two consecutive convolutional layers.\\
The primary objective of the pooling layer is to mitigate the risk of overfitting by reducing the spatial resolution of the feature map. By subsampling the feature map, the pooling layer effectively decreases the number of parameters and computational complexity within the model. This reduction in resolution aids in capturing the most significant information while discarding less relevant details. It helps in improving the model's generalization ability and preventing excessive reliance on specific spatial locations. The pooling function is denoted as:
\begin{equation}
	y_{i,j,k}^{l}=pool(z_{m,n,k}^{l}),\forall(m,n)\in{\mathcal R}_{i,j}
\end{equation}
Where $R_(i,j)$ is a local neighbourhood around position $(i,j)$. Each feature map in the pooling layer is connected to the corresponding feature map of the previous convolutional layer. There are several forms of pooling operations, the most common of which is maximum pooling. Maximum pooling layers use a $2\times2$ sized kernel to pool the input along the spatial dimension in steps of 2. Such an operation reduces the activation map to 25\% of its original size while maintaining the standard size of the deep convolution.\\
Additionally, the pooling layer can take the form of general pooling, in which the pooled neurons perform a variety of common operations, such as L1/L2 normalization and average pooling. However, in most convolutional neural networks, the most common form remains the maximum pooling layer.

\subsubsection{Fully-connected layer}

In CNNs, the fully connected layer typically resides between the convolutional layers and the output layer. Its purpose is to perform high-level inference and combine features. Each neuron in the fully connected layer is connected to all neurons in the preceding layer, enabling the formation of global semantic information. However, it's important to note that a fully connected layer is not always necessary, and a global average pooling layer can be used as an alternative. The global average pooling operation plays a role in reducing the spatial dimensionality of each feature map to 1 by pooling the entire feature map. This results in a global feature representation, which can be utilized effectively in various tasks.\\
In classification tasks, the fully connected layer is commonly employed as the output layer. It is often combined with a softmax operation or a Support Vector Machine (SVM) to merge the features extracted by the CNN with the best parameters specific to the given task. The optimization of these parameters can be achieved by minimizing an appropriate loss function defined for the particular task.\\
Suppose there are N desired input-output relationships, where $x(n)$ represents the nth input data, $y(n)$ represents the corresponding target label, and $o(n)$ represents the output of the CNN. The loss of the CNN can be calculated as follows:
\begin{equation}
	{\mathcal L}=\frac{1}{N}\sum_{n=1}^{N}\ell(\theta;y^{(n)},o^{(n)})
\end{equation}
By minimizing the loss function, the best fitting set of parameters can be found. Stochastic gradient descent (SGD) is a common solution for optimizing CNN networks.

\subsubsection{GAN}
\begin{figure}[H]
	\centering
	\includegraphics[scale=0.3]{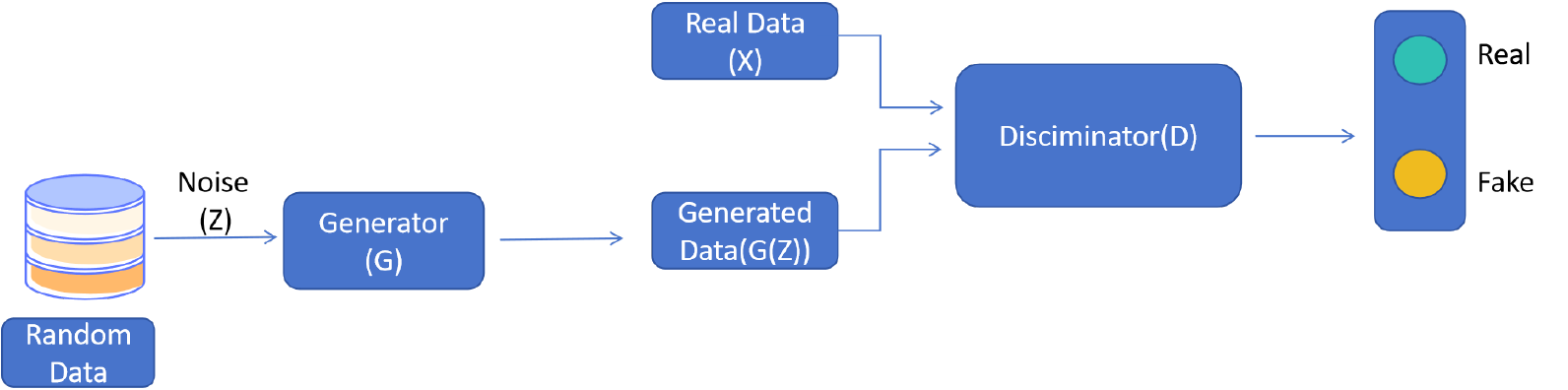}
	\caption{Basic Structure of Generative Adversarial Network(GAN).}
	\label{gan}
\end{figure}

The field of Artificial Intelligence (AI) has witnessed rapid advancements, leading to exponential growth in machines' ability to replicate human characteristics. A primary approach for simulating genuine human behavior involves the development of machine learning algorithms. An influential algorithm in this domain is Generative Adversarial Networks (GAN), introduced by Goodfellow in 2014\cite{NIPS2014_5ca3e9b1}. GANs operate on a fundamental principle inspired by a two-player zero-sum game, where the total gains of the players always sum up to zero. In this dynamic, each player's gain or loss of utility is precisely balanced by the corresponding loss or gain of the other player's utility.
GAN comprises two distinct models: the Generator (G) and the Discriminator (D), both trained through adversarial learning techniques. The Generative model (G) is tasked with generating novel input samples that are entirely unique from real input data. Despite lacking prior knowledge of the real input data, the Generator is designed to produce samples resembling the structure of authentic datasets. It takes random noise as input and generates new samples based on the distribution of real data samples.\\
On the other hand, the Discriminator (D) plays a pivotal role in distinguishing between real data and estimating the differences associated with data originating from the original dataset or the Generator. In Fig. \ref{gan}, 'Z' represents the noise vector, 'G(Z)' denotes the generated data, 'X' signifies the real data, 'G' represents the Generator, and 'D' stands for the Discriminator.
In GAN, both models, denoted as $(G, D)$, are engaged in a continual process of mutual learning, a topic we will delve into in the next section. The GAN model is represented by individual neural networks, with each network operating in opposition to the other. Specifically, $x$ is a sample drawn from the real data distribution $P_{data}(x)$, while $z$ is sampled from a prior distribution $p_z(z)$, such as a uniform or Gaussian distribution. The symbol E(·) represents the expectation operator. $D(x)$ represents the probability of $x$ being sampled from the real data rather than being generated. When the input data is from real data, the discriminator aims to maximize $D(x)$, approaching the value of 1. Conversely, when the input data is generated by the generator $G(z)$, the discriminator endeavors to make $D(G(z))$ approach 0 and generator $G$ tries to make it to 1.\\
Given the competitive nature of this zero-sum game involving two players, the optimization problem of GAN can be formulated as a minimax problem or the lossing function:

\begin{equation}
	\resizebox{0.45\textwidth}{!}{$
		\min_G\max_DV(D,G)=\mathbb{E}_{\boldsymbol{x}\sim p_{\mathrm{dan}}(\boldsymbol{x})}[\log D(\boldsymbol{x})]+\mathbb{E}_{\boldsymbol{z}\sim p_{\boldsymbol{z}}(\boldsymbol{z})}[\log(1-D(G(\boldsymbol{z}))]
		$}
\end{equation}
In summary, when training Generative Adversarial Networks (GANs), our objective is to learn the model's parameters through competitive training of the generator and discriminator. The generator's task is to produce realistic data, making it difficult for the discriminator to distinguish, while the discriminator aims to accurately differentiate between real and generated data. During the training process, the generator attempts to deceive the discriminator, while the discriminator strives to enhance its ability to discriminate between real and generated data. Through this competition, the generator and discriminator of GANs gradually improve their performance, ultimately achieving the goal of generating high-quality and realistic data.

\subsection{Experiment Design}
In our experiment, we employ various prominent architectures including Long Short-Term Memory (LSTM), Convolutional Neural Network (CNN), Generative Adversarial Network (GAN), as well as a hybrid model constructed by integrating LSTM and CNN. These models are selected to meticulously analyze and compare their capabilities in handling the specific challenges posed by our dataset and to meet the experiment objectives. Addtionally, to rigorously evaluate the impact of sentiment analysis on stock prediction performance, we designed a comprehensive set of experiments. These experiments consider both the content of stock-related comments and their associated topics across various computational models.

\subsubsection{Baseline (No Sentiment Analysis)}

In the baseline experiments, we assess the performance of various models without incorporating sentiment derived from comments or their topics.

\begin{itemize}
	\item \textbf{LSTM}: Utilizing the Long Short-Term Memory (LSTM) networks.
	\item \textbf{CNN}: Employing the Convolutional Neural Networks (CNN).
	\item \textbf{LSTM + CNN}: A hybrid approach that combines the strengths of both LSTM and CNN architectures.
	\item \textbf{GAN}: Leveraging the Generative Adversarial Networks (GAN) for prediction.
\end{itemize}

\subsubsection{Sentiment Analysis}

 For a more nuanced understanding, we further delve into the performance of models when sentiment analysis is incorporated. This sentiment is derived from both comments and their topics using two prominent sentiment analysis techniques: BERT and VADER.

\paragraph{BERT Sentiment Analysis}
Using the BERT model to extract sentiment. And The control experiment settings are as follows:

\begin{itemize}
	\item \textbf{LSTM}: 
	\begin{itemize}
		\item Without topic consideration.
		\item With topic consideration.
	\end{itemize}
	\item \textbf{CNN}: 
	\begin{itemize}
		\item Without topic consideration.
		\item With topic consideration.
	\end{itemize}
	\item \textbf{LSTM + CNN}: 
	\begin{itemize}
		\item Without topic consideration.
		\item With topic consideration.
	\end{itemize}
	\item \textbf{GAN}: 
	\begin{itemize}
		\item Without topic consideration.
		\item With topic consideration.
	\end{itemize}
\end{itemize}

Through this structured experimental design, we aim to provide a holistic understanding of the role of sentiment analysis in stock prediction, especially when juxtaposed against various computational models.

\begin{itemize}
	\item \textbf{LSTM}: 
	\begin{itemize}
		\item Without topic consideration.
		\\ i. Data Scaling: \\
		\hspace*{2em} To get sentiment (polarity) scores, we use VADER (Valence Aware Dictionary for Sentiment Reasoning) model.First, we will filter the comments to only those pertaining to Amazon stock. We will iterate through each line, preprocessing the comments with regularization. Using NLTK packages, we will categorize each comment as ``Positive'', ``Neutral'', or ``Negative'' and calculate an overall sentiment score. We will take a weighted arithmetic mean of the sentiment scores for comments on the same day to obtain a daily sentiment score.After that,We employed a dual-scaling approach to normalize our data without topic. First, the ``Adj Close'' column, representing the adjusted closing prices, was scaled using the MinMaxScaler. This scaler was applied to transform the ``Adj Close'' values into a specified range, maintaining the structure of the dataset while normalizing the values. Subsequently, we scaled the entire dataset, again utilizing the MinMaxScaler, to ensure that all features contributed equally to the analysis.
		\\ii. Dataset Preparation for train and test\\
		\hspace*{2em}After data scaling,we divided our data into training and testing sets. The training set comprised all data points except the last 20 rows, whereas the testing set consisted of the final 20 rows. To accommodate the LSTM model, we prepared the data in a specific format. This involved creating datasets with a specified batch size, ensuring that each input sample had an associated target value from the ``Adj Close'' column.
		\\iii. LSTM Model Development \\
		\hspace*{2em}Our model was built using a Sequential approach with Keras. And we input the 15 variable except date to predict the Adj Close.The core of the model was an LSTM layer with 50 units and a ``relu'' activation function. This LSTM layer was designed to process input data in the shape of 3D arrays, with dimensions corresponding to samples, time steps, and features. The model concluded with a Dense layer consisting of a single unit for output. For compiling the model, we used the Adam optimizer and mean squared error (MSE) as the loss function.
		\\iv. Training and Predictions\\
		\hspace*{2em}The model was trained on the prepared training set for 200 epochs. Training was conducted silently, without outputting progress logs. Post-training, we employed the model to make predictions on both the training and testing datasets.
		\\v. Outcome Analysis \\
		\hspace*{2em}We generated three figures - predicted images for the training set Fig.\ref{fig:11}, predicted images for the test set Fig.\ref{fig:12}, and an image of the loss function for the LSTM model Fig. \ref{fig:13} Additionally, we calculated the RMSE, MAE, R-squared, and MAPE for both the training and test sets, as shown in Table XX. Examining the loss function, we see that it converged quickly, reaching stability after approximately 30 epochs. The LSTM model performed reasonably well, capturing the tendency for historical stock price data to diminish in influence over time in financial markets. However, there is still room for improvement in predictive performance. Because our sentiment scores don't cover the topic,we will try to use topic sentiment for improving our model.
		\begin{figure}[h]
			\begin{minipage}[t]{0.48\textwidth}
				\centering
				\includegraphics[width=\linewidth]{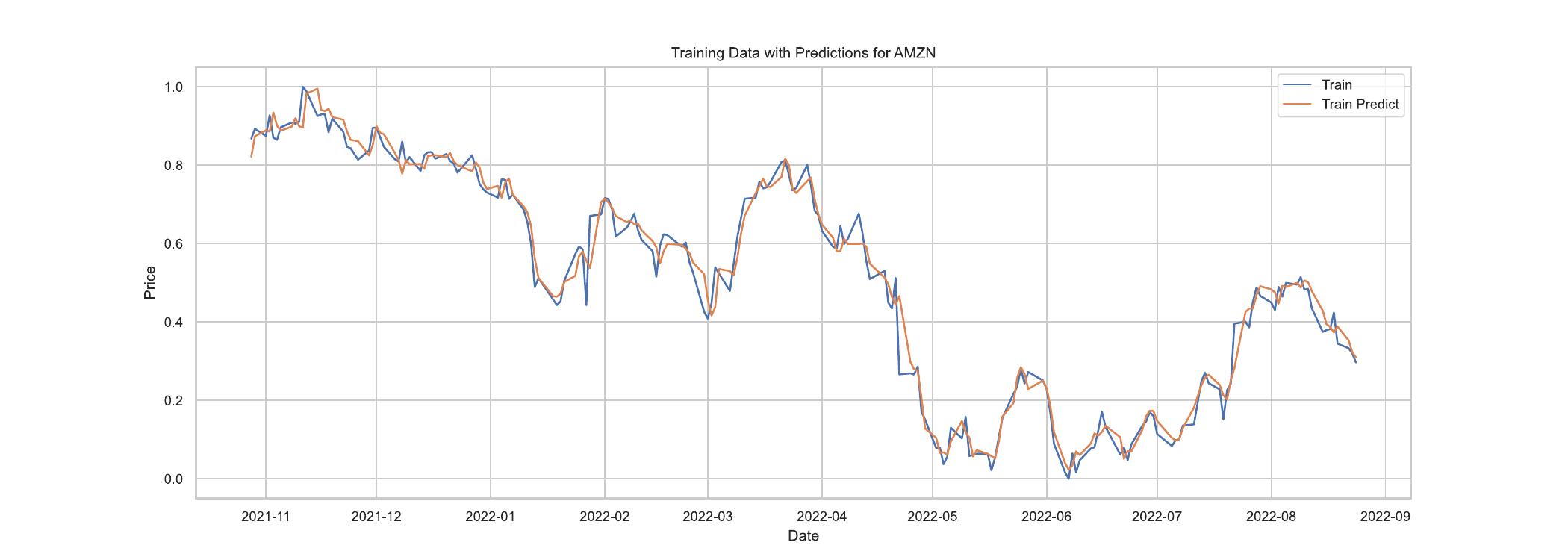}
				\caption{Training for AMZN (LSTM)}
				\label{fig:11}
				\vspace{\floatsep} 
				\includegraphics[width=\linewidth]{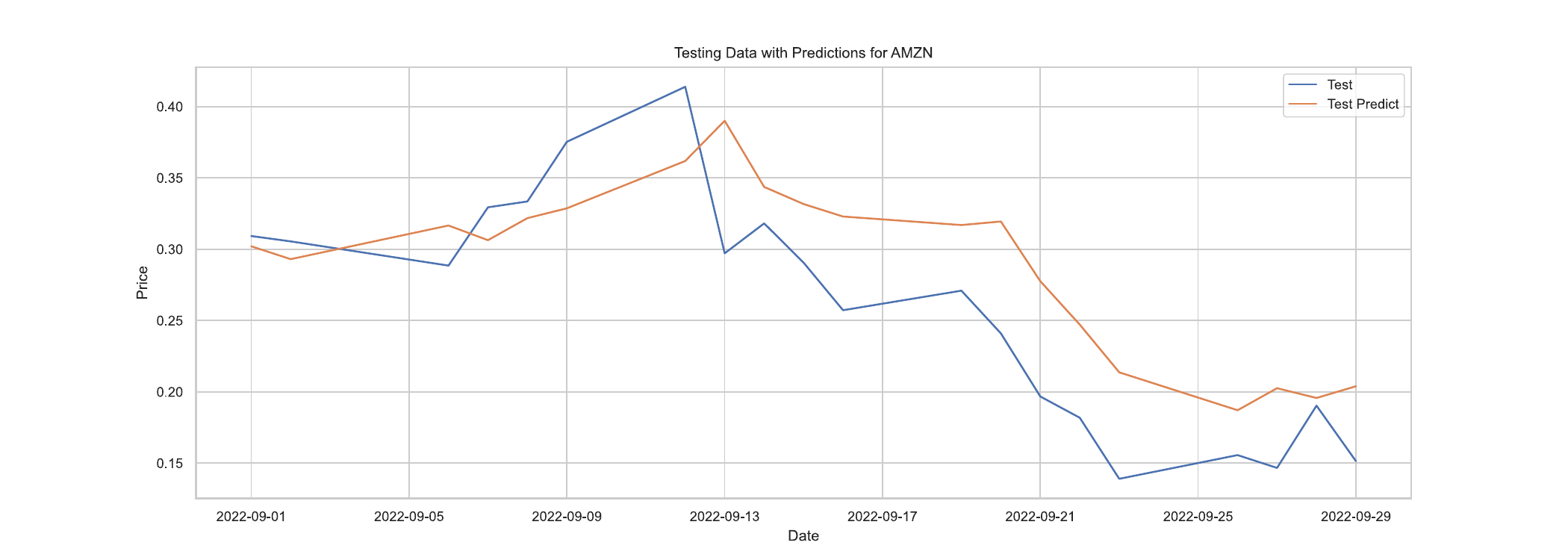}
				\caption{Testing for AMZN (LSTM)}
				\label{fig:12}
			\end{minipage}
			\hfill 
			\begin{minipage}[t]{0.48\textwidth}
				\centering
				\includegraphics[width=\linewidth]{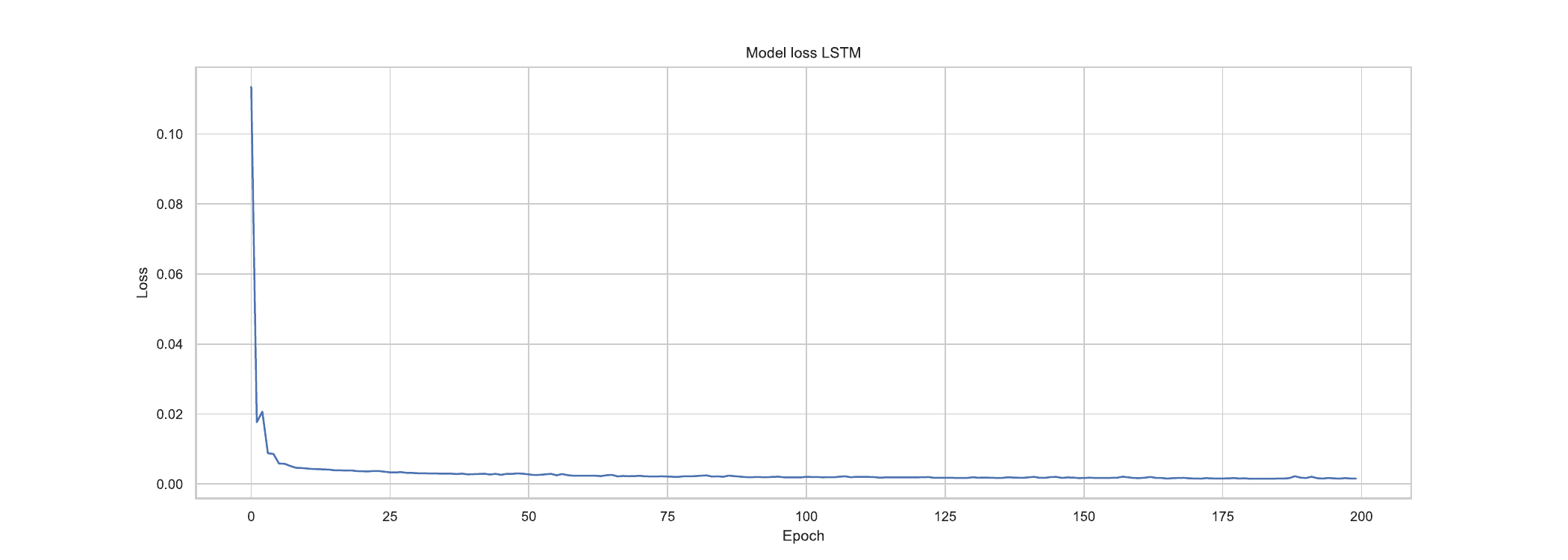}
				\caption{Training loss of LSTM}
				\label{fig:13}
			\end{minipage}
		\end{figure}
		\begin{table}[h]
			\centering
			\caption{Performance Evaluation Metrics}
			\label{TABLE:III}
			\begin{tabular}{|c|c|c|}
				\hline
				TITLE  & LSTM & LSTM(Topic) \\ 
				\hline
				Train Set   &   &\\ 
				\hline
				RMSE   &  3.122 & 3.726\\
				MAE  & 2.279  & 2.703\\
				R2 Score  & 0.982  & 0.972\\
				MAPE  & 1.166  & 1.972\\
				\hline
				Test Set   &   &\\ 
				\hline
				RMSE   &  4.261 & \textbf{3.802}\\
				MAE  & 3.699  & \textbf{3.026}\\
				R2 Score  & 0.569  & \textbf{0.657}\\
				MAPE  & 3.025  & \textbf{2.432}\\
				\hline
			\end{tabular}
			
		\end{table}
		
		\item With topic consideration
		\\While keeping the preceding steps consistent, we incorporated the bert library to transform the previously sentiment-only scores into topic-aware scores. This allowed us to integrate topical information into the overall model training process. Specifically, the solely sentiment-based scores were augmented with topics using bert, and these enhanced scores were then fed into the model for training.Similarly, we generated three figures and one table.By incorporating topic modeling using bert, the performance of the model improved markedly. Quantitatively, the key evaluation metrics of RMSE, MAE, R-squared, and MAPE improved by about 10\%
		\begin{figure}[h]
			\begin{minipage}[t]{0.48\textwidth}
				\centering
				\includegraphics[width=\linewidth]{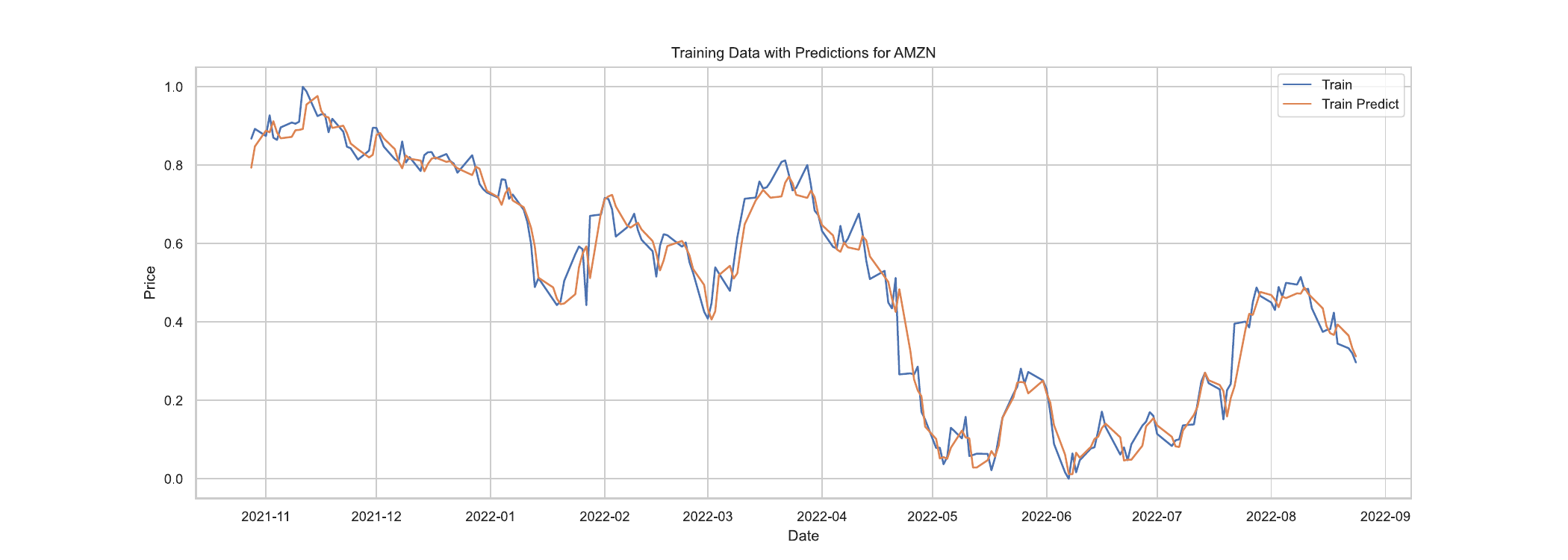}
				\caption{Training for AMZN (LSTM with Topic)}
				\label{fig:14}
				\vspace{\floatsep} 
				\includegraphics[width=\linewidth]{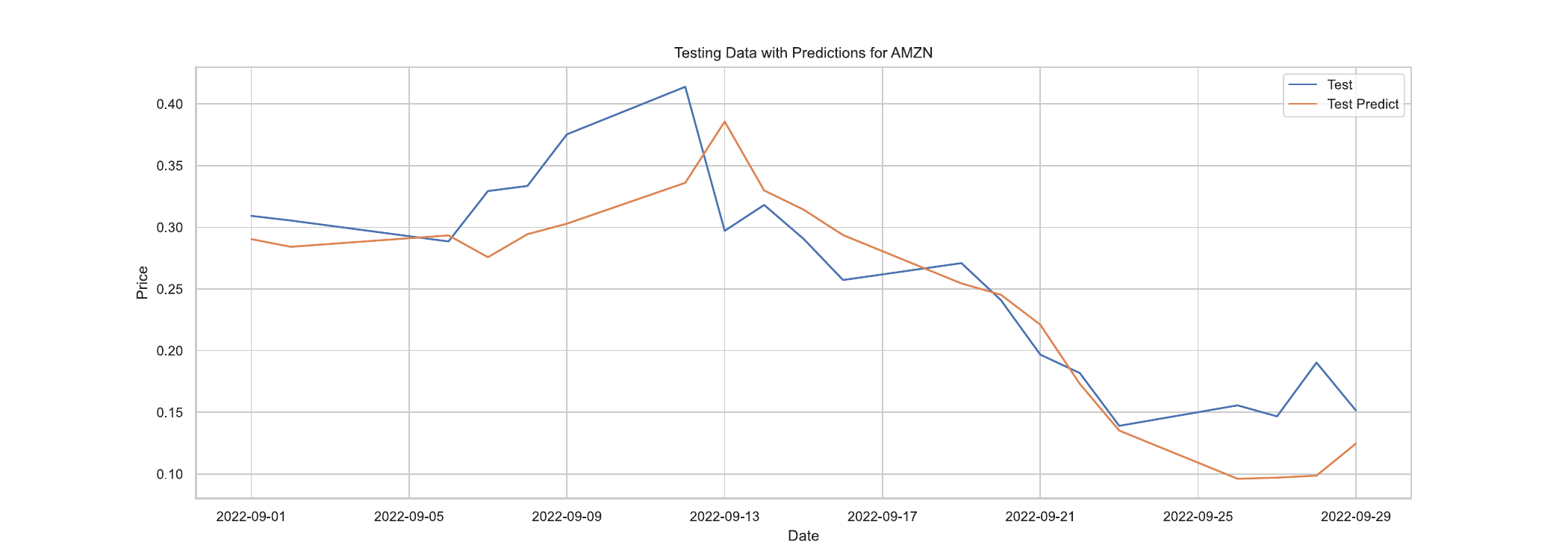}
				\caption{Testing for AMZN (LSTM with Topic)}
				\label{fig:15}
			\end{minipage}
			\hfill 
			\begin{minipage}[t]{0.48\textwidth}
				\centering
				\includegraphics[width=\linewidth]{loss_LSTM.pdf}
				\caption{Training loss of LSTM with topic}
				\label{fig:16}
			\end{minipage}
		\end{figure}
%
		
	\end{itemize}
	\item \textbf{CNN}: 
	\begin{itemize}
		\item Without topic consideration.
		\\i CNN Model Architecture and Training\\
		 \hspace*{2em}When employing the CNN model, the data preprocessing and training procedures were kept consistent with the prior CNN approach.The model architecture was designed using the Sequential API from TensorFlow Keras. The primary layer was a Conv1D layer with 64 filters and a kernel size of 2, employing the 'relu' activation function. This was followed by a MaxPooling1D layer with a pool size of 2 to reduce dimensionality and a Flatten layer to convert the pooled feature map into a 1D array. The network concluded with two Dense layers, the first with 50 neurons (also using 'relu' activation) and the final layer producing a single output value. The model was compiled using the Adam optimizer and mean squared error (MSE) as the loss function.
		\\ii Outcome Analysis\\
		\hspace*{2em}Since CNNs are an architecture better suited for computer vision tasks, they underperformed LSTMs for this time series forecasting application where temporal weights are influential. As with the LSTM experiments, we generated 3 figures to visualize model predictions and loss, along with a table summarizing performance on key evaluation metrics. The predictions made by the CNN model exhibited greater volatility and variance.
		\begin{figure}[h]
			\begin{minipage}[t]{0.48\textwidth}
				\centering
				\includegraphics[width=\linewidth]{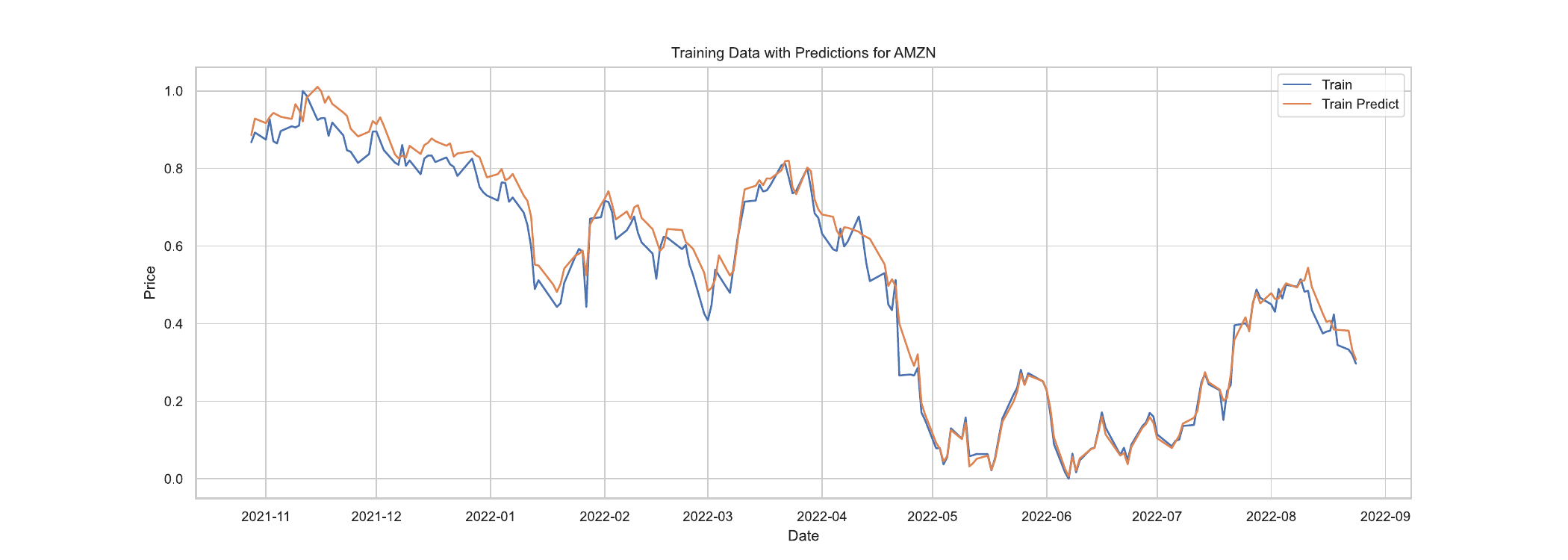}
				\caption{Training for AMZN (CNN)}
				\label{fig:17}
				\vspace{\floatsep} 
				\includegraphics[width=\linewidth]{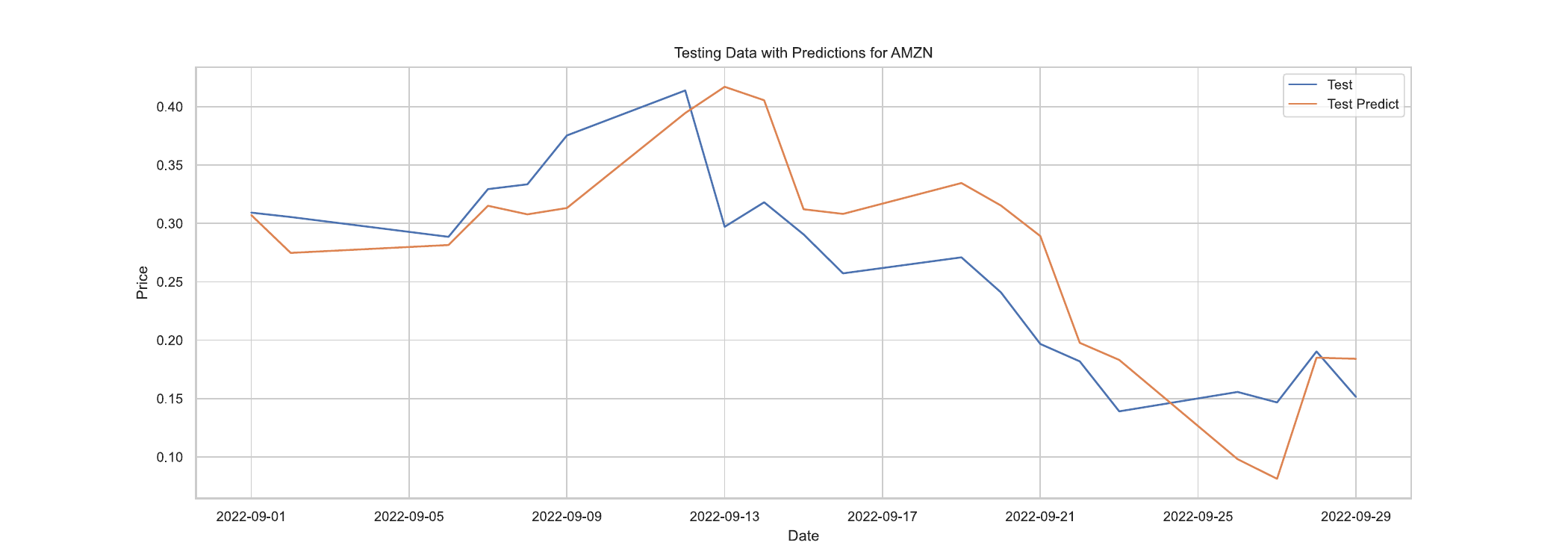}
				\caption{Testing for AMZN (CNN)}
				\label{fig:18}
			\end{minipage}
			\hfill 
			\begin{minipage}[t]{0.48\textwidth}
				\centering
				\includegraphics[width=\linewidth]{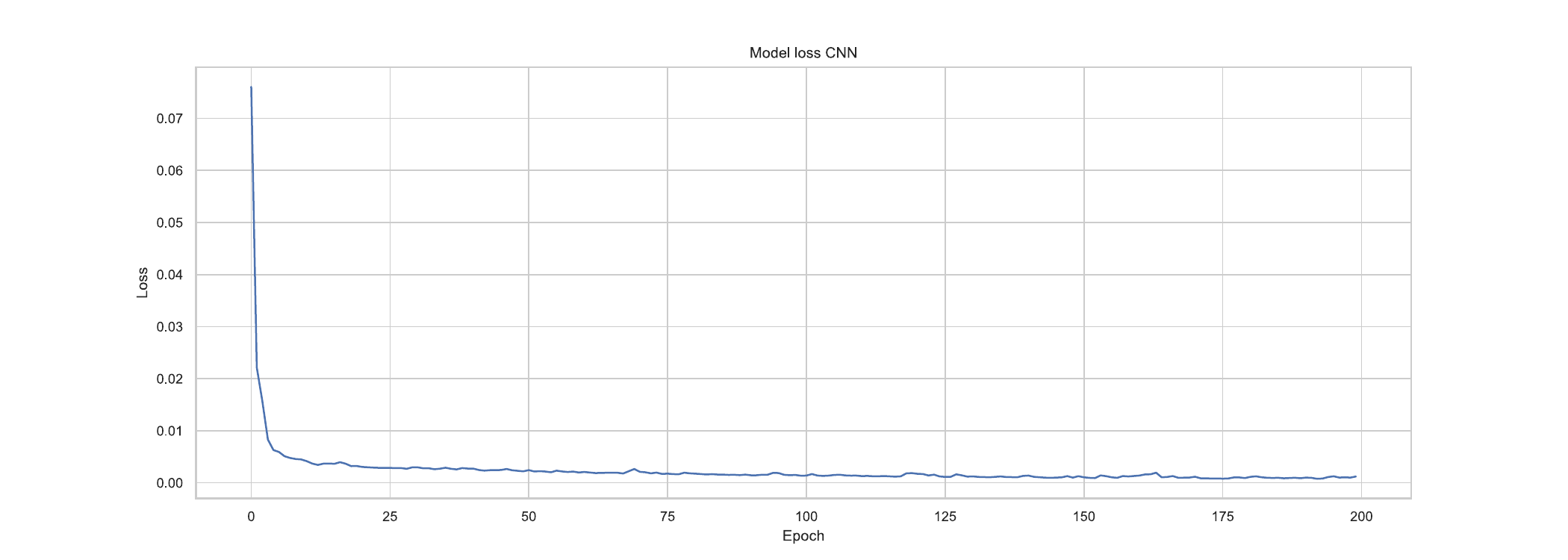}
				\caption{Training loss of CNN }
				\label{fig:19}
			\end{minipage}
		\end{figure}
		\begin{table}[h]
			\centering
			\caption{Performance Evaluation Metrics}
			\label{TABLE:IV}
			\begin{tabular}{|c|c|c|}
				\hline
				TITLE  & CNN & CNN(Topic)\\ \hline
				Train Set   &  & \\ 
				\hline
				RMSE   &  2.107 & 2.881\\
				MAE  & 1.421  & 2.143\\
				R2 Score  & 0.992  & 0.985\\
				MAPE  & 0.983  & 1.610\\
				\hline
				Test Set   &  & \\ 
				\hline
				RMSE   &  5.016 & \textbf{4.724}\\
				MAE  & 4.255  & \textbf{4.075}\\
				R2 Score  & 0.403  & \textbf{0.470}\\
				MAPE  & 3.409 & \textbf{3.323}\\
				\hline
			\end{tabular}
			
		\end{table}
		\\\\
		\item With topic consideration.
		\\ When incorporating topical sentiment scores while holding other factors constant, the CNN model exhibited noticeable improvements across all evaluation metrics.To visually and quantitatively illustrate the performance gains, comparative figures were constructed contrasting the CNNs with and without topical sentiment scores.\\
		\begin{figure}[h]
			\begin{minipage}[t]{0.48\textwidth}
				\centering
				\includegraphics[width=\linewidth]{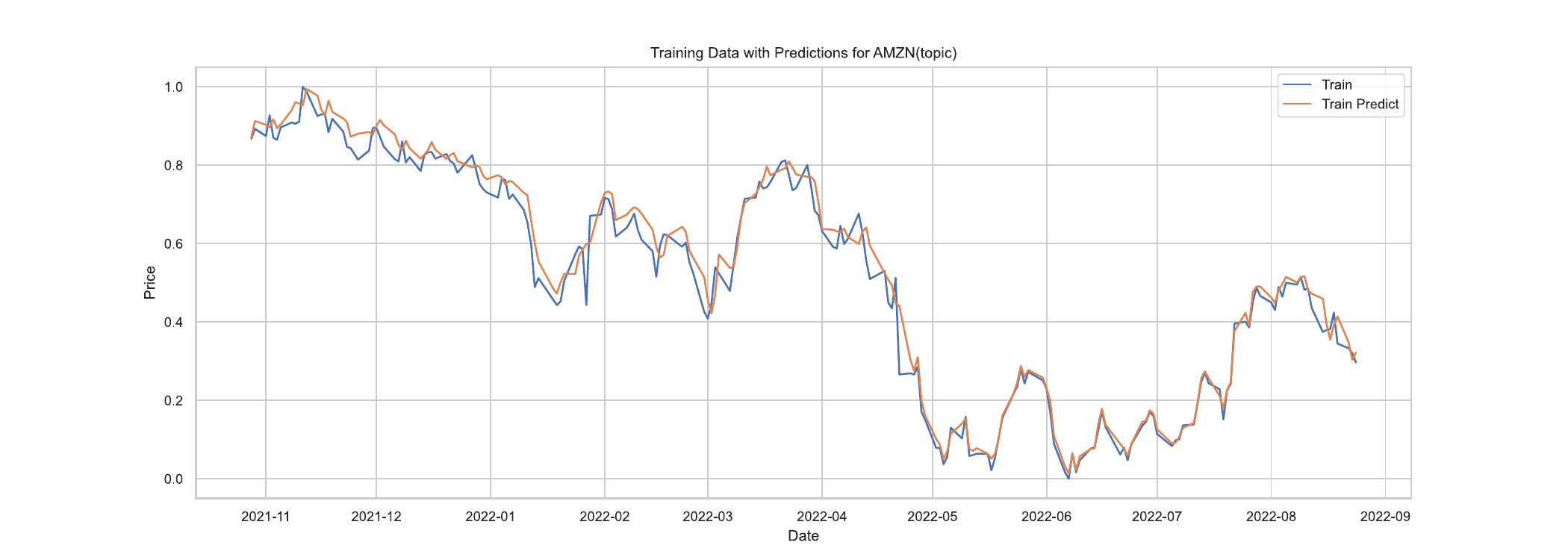}
				\caption{Training for AMZN (CNN with Topic)}
				\label{fig:20}
				\vspace{\floatsep} 
				\includegraphics[width=\linewidth]{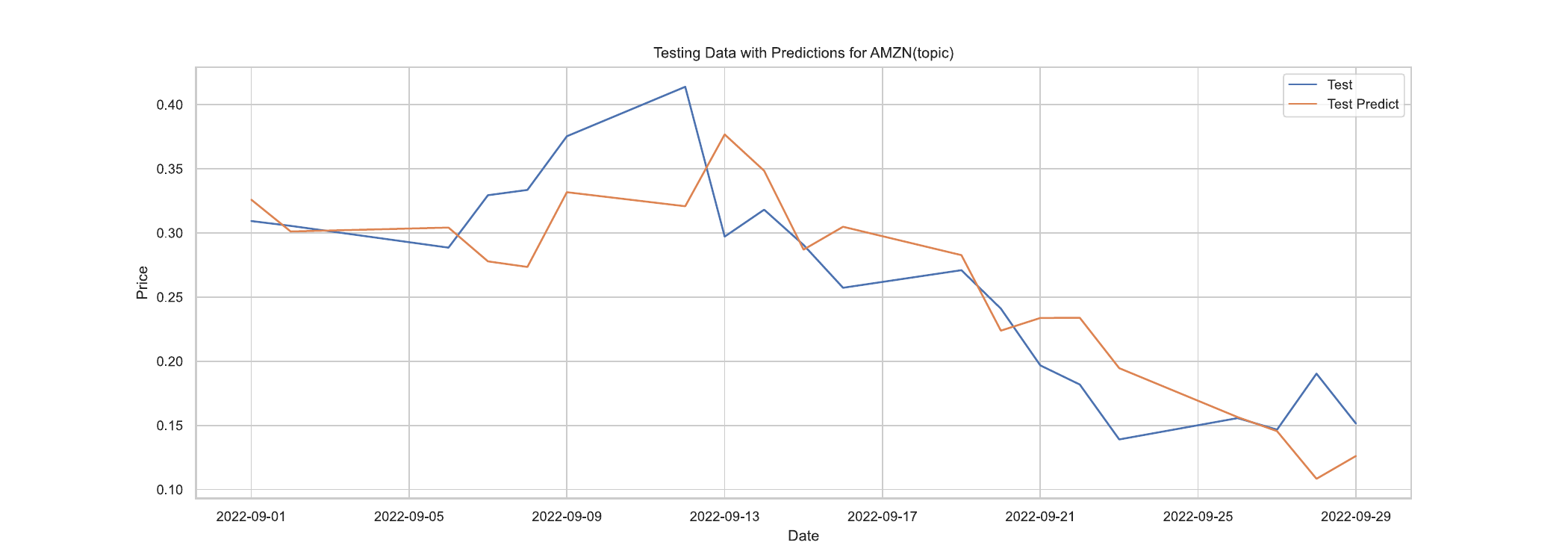}
				\caption{Testing for AMZN (CNN with Topic)}
				\label{fig:21}
			\end{minipage}
			\hfill 
			\begin{minipage}[t]{0.48\textwidth}
				\centering
				\includegraphics[width=\linewidth]{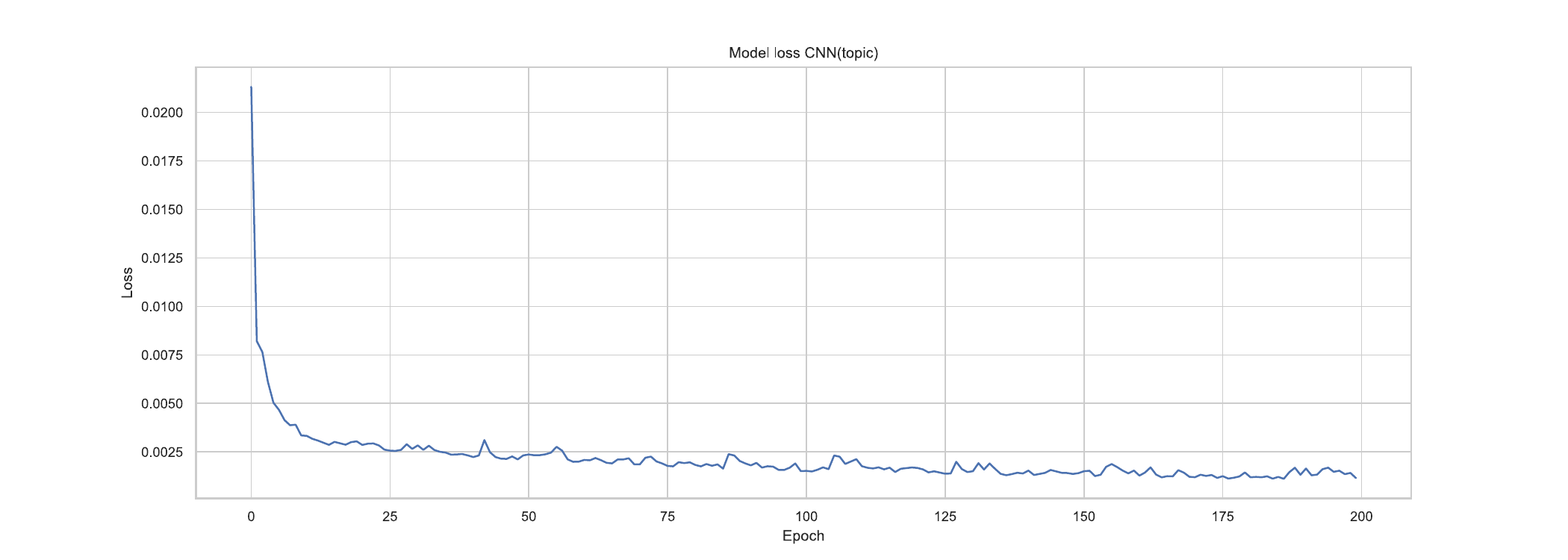}
					\caption{Training loss of CNN with Topic}
				\label{fig:22}
			\end{minipage}
		\end{figure}
%
		\\
	\end{itemize}
	\item \textbf{LSTM + CNN}: 
	\begin{itemize}
		\item Without topic consideration.\\
		i. Build the model \\
		\hspace*{2em}A CNN-LSTM model brings together the best of both worlds – the feature extraction prowess of CNNs and the sequential data processing strength of LSTMs. This synergy makes it a potent tool for tackling complex problems in sequential and time-series data analysis.The model architecture combined a Convolutional Neural Network (CNN) and a Long Short-Term Memory (LSTM) network using the Sequential API from Keras. The primary layer, a Conv1D, had 64 filters and a kernel size of 2 with ``relu'' activation, designed for extracting spatial features from the input data. Following this, an LSTM layer with 50 neurons and 'relu' activation captured temporal dependencies. The network concluded with a Dense layer outputting a single value. The model was compiled using the Adam optimizer with mean squared error (MSE) as the loss function.
		\\ii outcome analysis\\
		\hspace*{2em}As hypothesized, combining the CNN and LSTM architectures via an ensemble model led to enhanced overall predictive performance:
		\begin{figure}[h]
			\begin{minipage}[t]{0.48\textwidth}
				\centering
				\includegraphics[width=\linewidth]{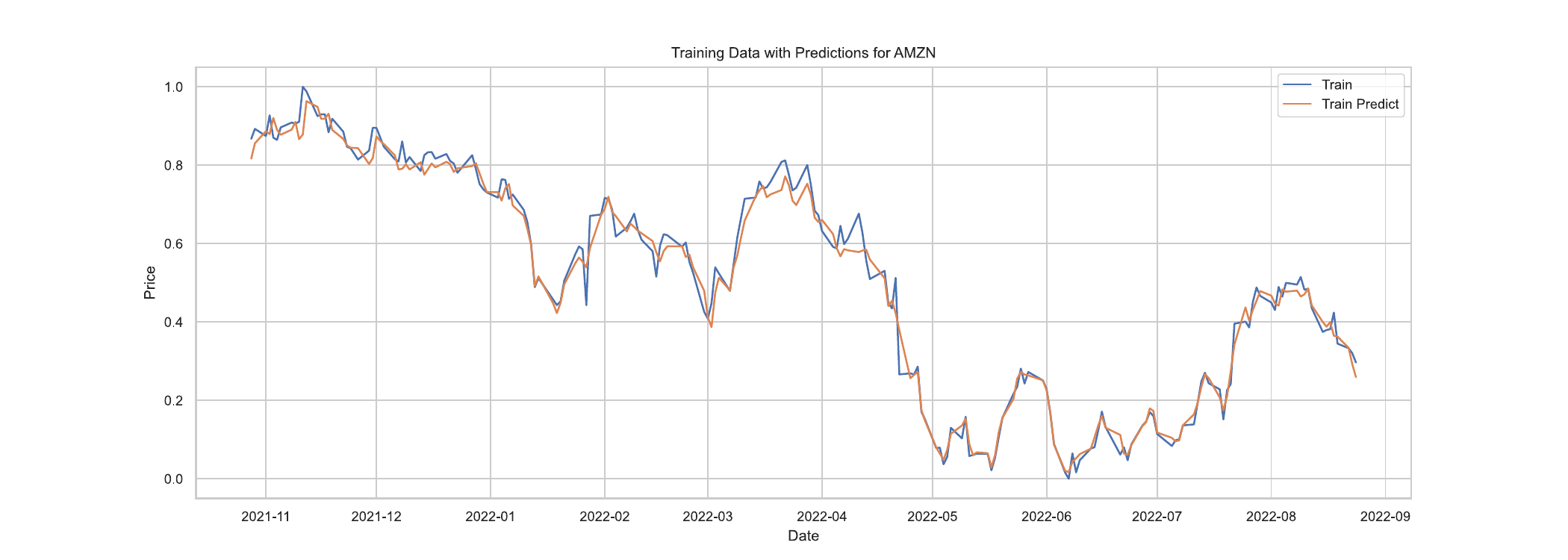}
				\caption{Training for AMZN (LSTM\&CNN)}
				\label{fig:23}
				\vspace{\floatsep} 
				\includegraphics[width=\linewidth]{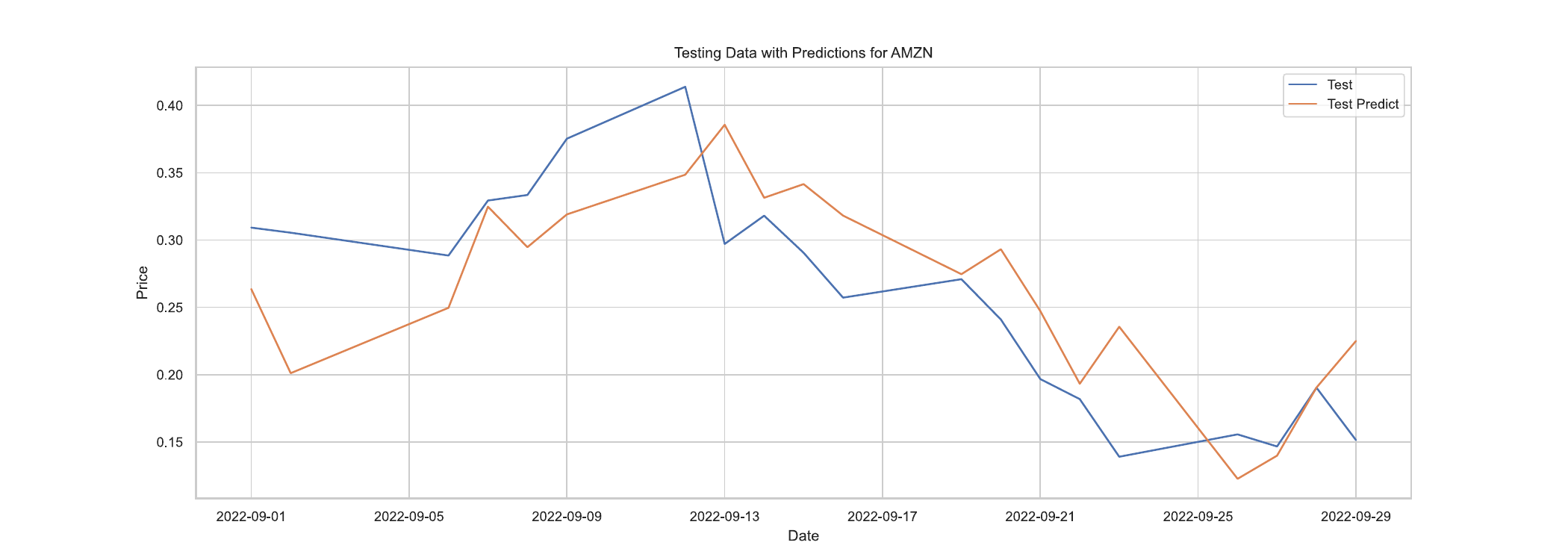}
				\caption{Testing for AMZN (LSTM\&CNN)}
				\label{fig:24}
			\end{minipage}
			\hfill 
			\begin{minipage}[t]{0.48\textwidth}
				\centering
				\includegraphics[width=\linewidth]{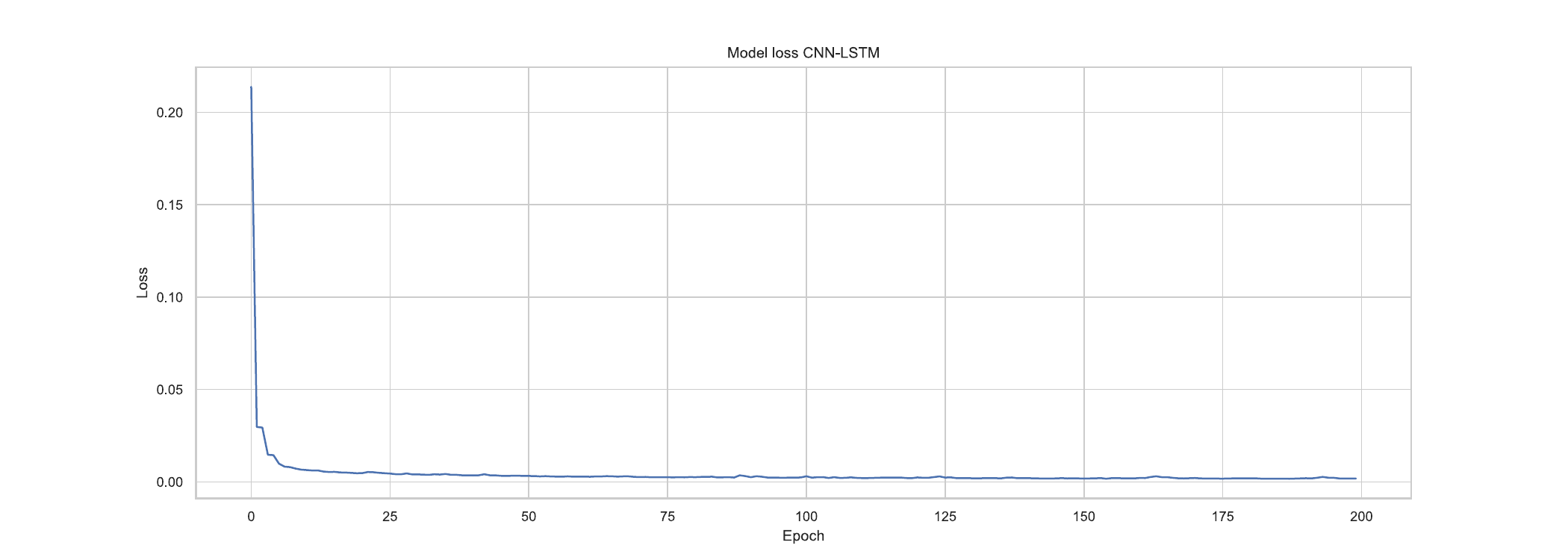}
				\caption{Training loss of LSTM\&CNN}
				\label{fig:25}
			\end{minipage}
		\end{figure}
		\begin{table}[h]
			\centering
			\caption{Performance Evaluation Metrics}
			\label{TABLE:V}
			\begin{tabular}{|c|c|c|}
				\hline
				TITLE  & CNN-LSTM & CNN-LSTM(Topic) \\ \hline
				Train Set   &   & \\ 
				\hline
				RMSE   &  2.640  & 3.533\\
				MAE  & 1.936  & 2.556\\
				R2 Score  & 0.987 & 0.977 \\
				MAPE  &  1.315 & 1.792\\
				\hline
				Test Set   &   &\\ 
				\hline
				RMSE   &  4.482 & \textbf{3.509}\\
				MAE  & 3.693 & \textbf{2.940}\\
				R2 Score  & 0.523  & \textbf{0.708}\\
				MAPE  & 2.984 & \textbf{2.360}\\
				\hline
			\end{tabular}
			
		\end{table}

		\item With topic consideration.
		\\ Certainly, in addition to using the topic as a control group, we also incorporated topics into our experiments using a CNN-LSTM model for overall stock price prediction. The data showed that our experiments with the combined CNN-LSTM model outperformed individual CNN and LSTM models containing topics. \\
		
		\begin{figure}[H]
			\begin{minipage}[t]{0.48\textwidth}
				\centering
				\includegraphics[width=\linewidth]{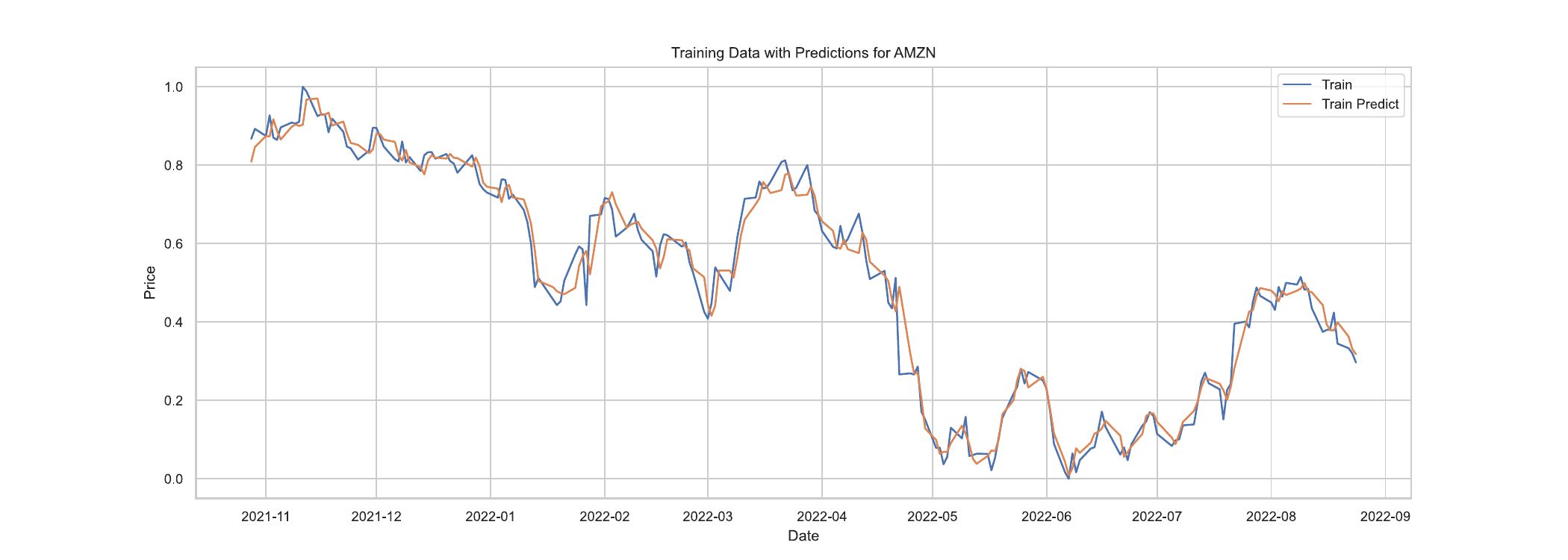}
				\caption{Training for AMZN (LSTM\&CNN with Topic) }
				\label{fig:26}
				\vspace{\floatsep} 
				\includegraphics[width=\linewidth]{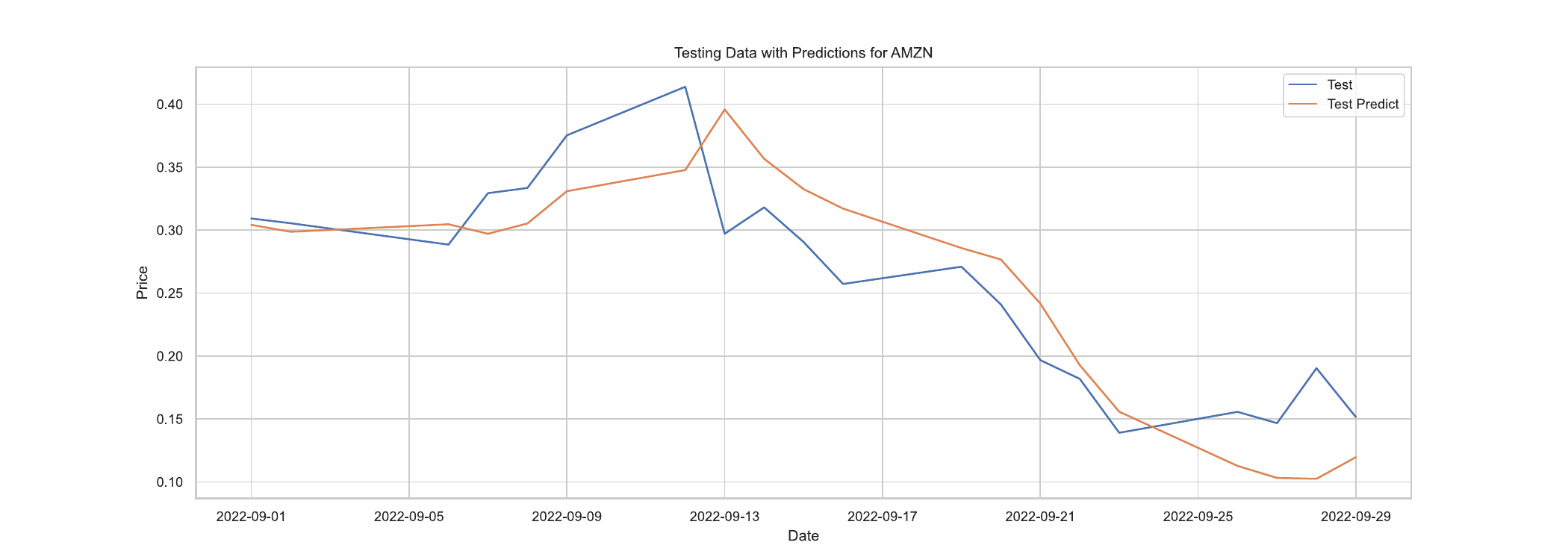}
				\caption{Testing for AMZN (LSTM\&CNN with Topic)}
				\label{fig:27}
			\end{minipage}
		\end{figure}
		\begin{figure}[H]
			\centering
			\includegraphics[width=\linewidth]{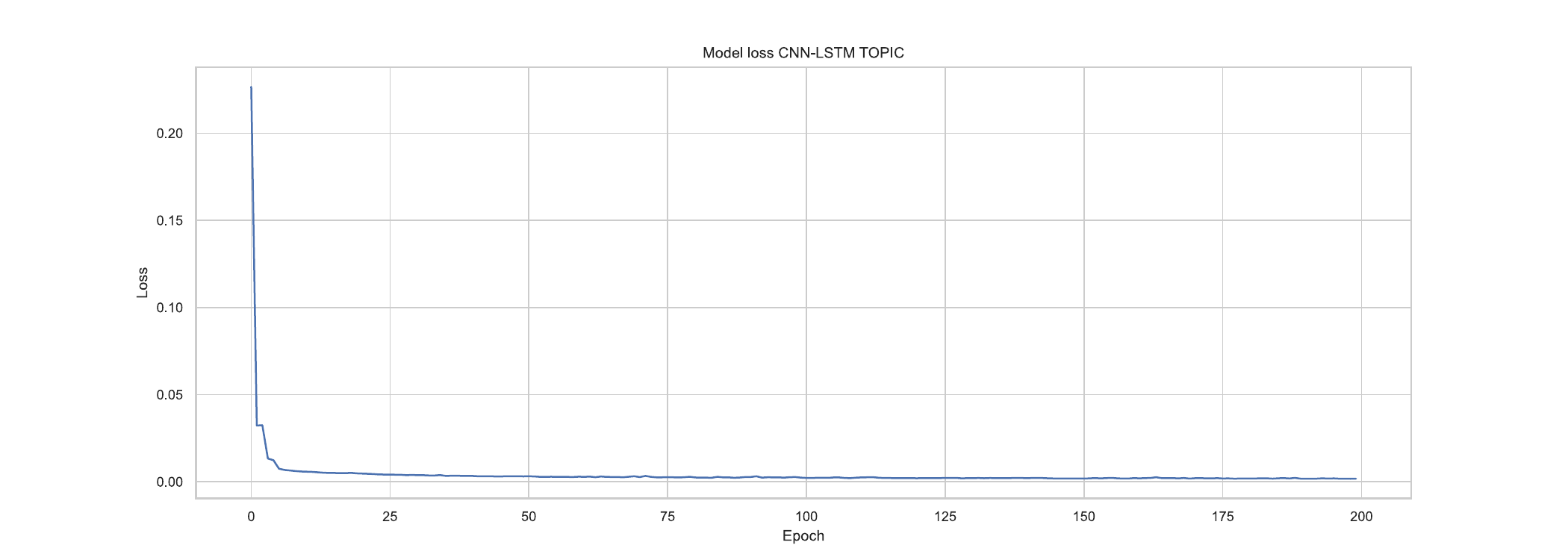}
			\caption{Training loss of LSTM\&CNN}
			\label{fig:28}
		\end{figure}
%
	\end{itemize}

	\item \textbf{GAN}: 
	\begin{itemize}
		\item Without topic consideration.
		\begin{figure}[H]
			\centering
			\includegraphics[width=\linewidth]{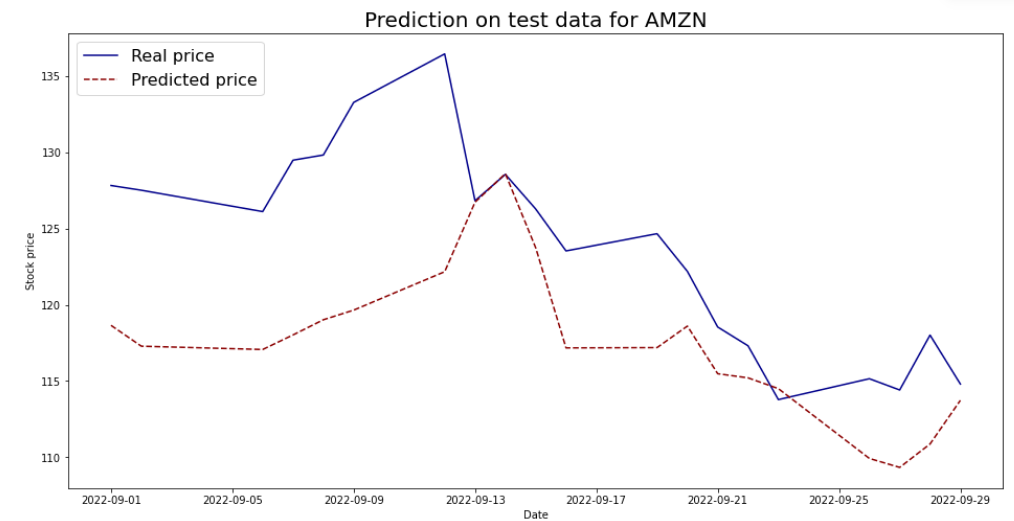}
			\caption{Testing for AMZN (GAN)}
			\label{fig:29}
		\end{figure}
		\item With topic consideration.
		\begin{figure}[H]
			\centering
			\includegraphics[width=\linewidth]{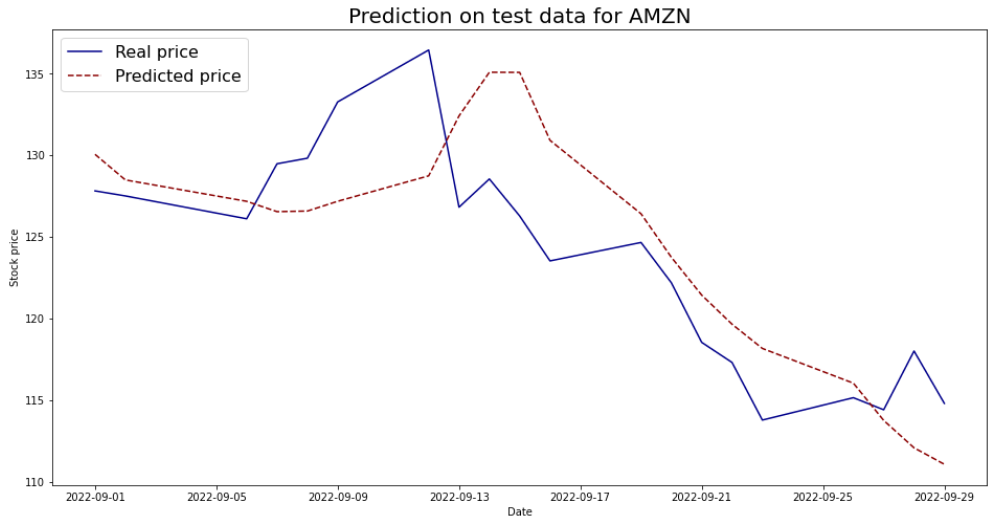}
			\caption{Testing for AMZN (GAN with Topic)}
			\label{fig:30}
		\end{figure}
	\end{itemize}
			\begin{table}[h]
		\centering
		\caption{Performance Evaluation Metrics}
		\label{TABLE:VI}
		\begin{tabular}{|c|c|c|}
			\hline
			TITLE  & GAN & GAN(Topic) \\ \hline
			Test Set   &   & \\ 
			\hline
			RMSE   & 7.560   & \textbf{4.569}\\
			MAE  & 6.155  & \textbf{3.835}\\
			R2 Score  & -1.859 & \textbf{0.558} \\
			MAPE  &  5.2465 & \textbf{3.046}\\
			\hline
		\end{tabular}
		
	\end{table}
\end{itemize}

\begin{figure}[H]
	\begin{minipage}[t]{0.48\textwidth}
		\centering
		\includegraphics[width=\linewidth]{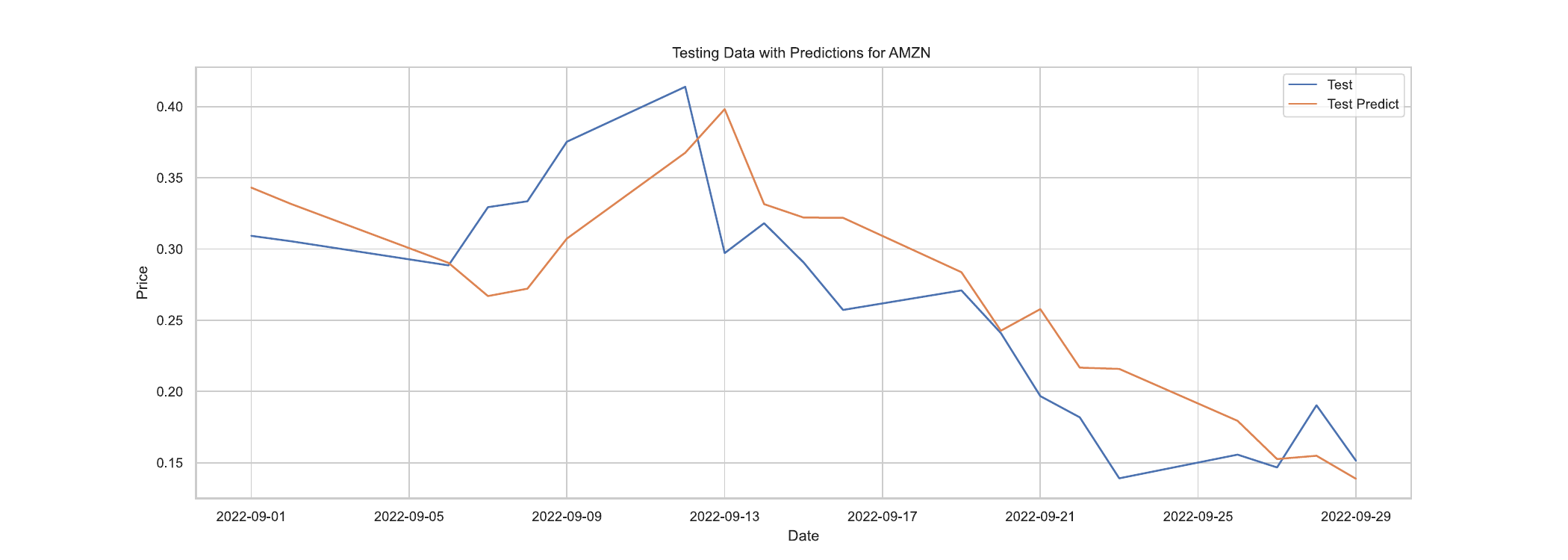}
		\caption{Test for AMZN (CNN with VADER sentiment) }
		\label{fig:31}
		\vspace{\floatsep} 
		\includegraphics[width=\linewidth]{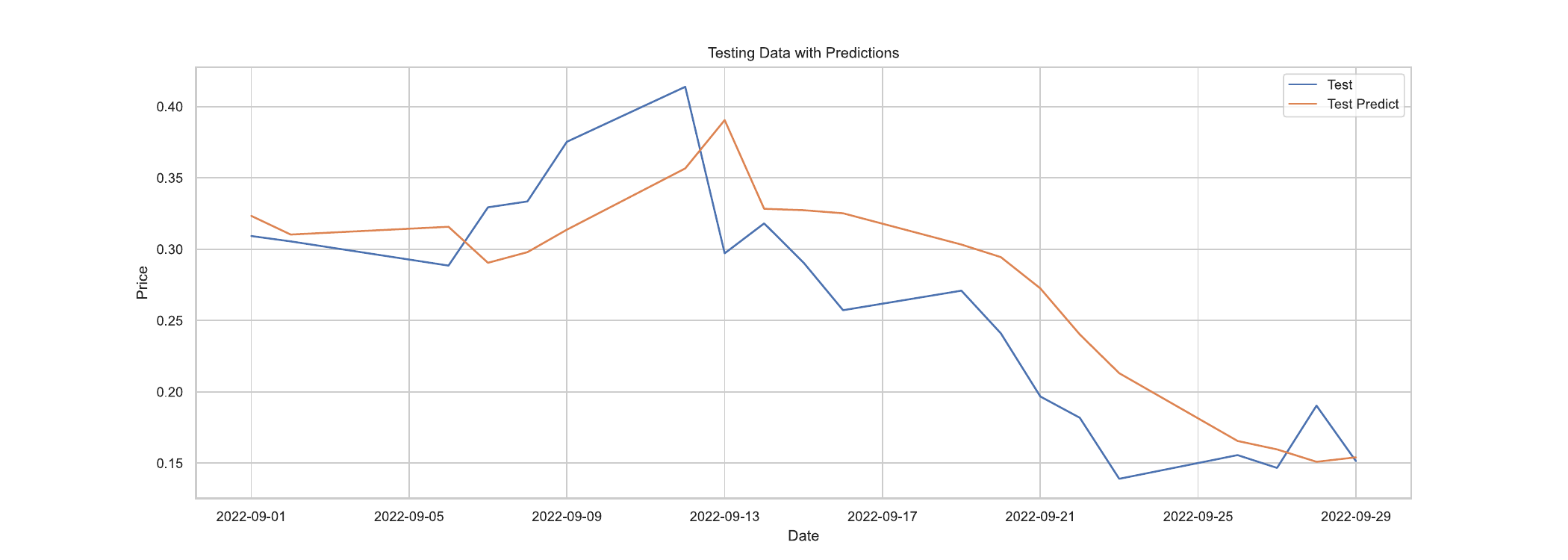}
		\caption{Testing for AMZN (LSTM with VADER sentiment)}
		\label{fig:32}
	\end{minipage}
\end{figure}

\begin{figure}[H]
	\centering
	\includegraphics[width=\linewidth]{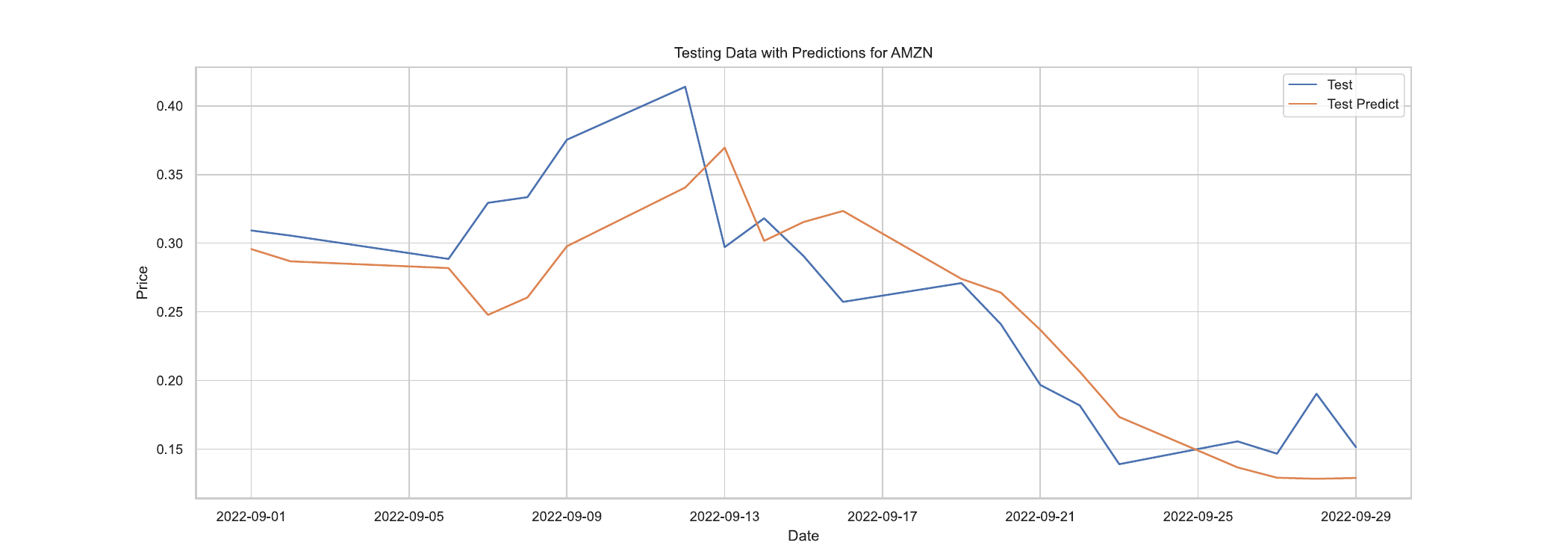}
	\caption{Testing for AMZN (LSTM\&CNN with VADER sentiment)}
	\label{fig:33}
\end{figure}

\paragraph{VADER Sentiment Analysis}
We utilize the VADER tool for sentiment extraction, maintaining a consistent experimental setup with the one described in the \textit{BERT Sentiment Analysis} section. Consequently, we will present only the resultant tables and charts here, as the experimental parameters remain unchanged. We visualized the results using graphical representations Figs. \ref{fig:31}, \ref{fig:32}, \ref{fig:33} and Table \ref{TABLE:VII}. Displayed from top to bottom are the test set fitting effects of CNN, LSTM, and a CNN-LSTM hybrid model, respectively. 
	\begin{table}[htbp]
	\centering
	\caption{Performance Evaluation Metrics}
	\label{TABLE:VII}
	\resizebox{0.45\textwidth}{!}{
		\begin{tabular}{|c|c|c|c|c|c|c|}
			\hline
			TITLE  & CNN & CNN(Vader\&TOPIC) & LSTM & LSTM(Vaderr\&TOPIC) & CNN-LSTM & CNN-LSTM(Vaderr\&TOPIC)\\ \hline
			Test Set   &   &  &  &   &  &\\ 
			\hline
			RMSE   & 5.016 & \textbf{3.999} & 4.261 & \textbf{4.151} & 4.482 & \textbf{3.844}\\
			MAE  & 4.255 & \textbf{3.583} & 3.699 & \textbf{3.579}  & 3.693 & \textbf{3.177}\\
			R2 Score  & 0.403 & \textbf{0.620} & 0.569 & \textbf{0.591} & 0.523 & \textbf{0.649} \\
			MAPE  & 3.409 &  \textbf{2.872} & 3.025 & \textbf{2.915} & 2.984 & \textbf{2.537}\\
			\hline
		\end{tabular}
	}
\end{table}
As indicated in Table \ref{TABLE:VII}, employing VADER as the sentiment analysis tool for scoring the topic feature yields improved performance.

Overall, across the three experimental groups, sentiment analysis of topics generated with BERT led to markedly better stock price prediction compared to models without topics. We visualized the results using graphical representations Figs. \ref{fig:21}, \ref{fig:24}, \ref{fig:27}, \ref{fig:30}, \ref{fig:31}, \ref{fig:32} and \ref{fig:33}  and Tables \ref{TABLE:III}, \ref{TABLE:IV}, \ref{TABLE:V}, \ref{TABLE:VI}, and \ref{TABLE:VII}. When the topic feature is integrated into the predictive model, there is a notable improvement in performance metrics: the Root Mean Squared Error (RMSE), Mean Absolute Error (MAE), and Mean Absolute Percentage Error (MAPE) all decrease compared to the variant of the model that excludes the topic feature. Concurrently, there is an appreciable increase in the R-squared (R2) score.\\
These enhancements in the metrics convey that the topic feature substantively contributes to the model's explanatory power. A diminished RMSE reflects a tighter clustering of the model's predicted values around the actual data points, suggesting increased prediction accuracy. Similarly, reductions in MAE and MAPE indicate a higher precision of the model's forecasts. The elevated R2 score reveals that the model with the topic feature accounts for a greater proportion of the variance in the dependent variable, signifying a more robust model fit. In essence, the addition of the topic feature substantially bolsters the model's predictive validity and its capacity to elucidate the underlying patterns in the dataset.

\section*{Conclusion and Discussion}
In this paper, we have undertaken a detailed evaluation of the role of sentiment analysis, specifically the sentiment of topics extracted from stock market comments, in predicting stock prices. Utilizing BERTopic, a sophisticated Natural Language Processing (NLP) technique, as our primary tool for topic modeling, we have ventured into the realm of advanced analytics in the stock market domain. Our approach has not been limited to traditional methods; rather, it has incorporated a variety of models that are acclaimed for their effectiveness in time series and stock prediction tasks. Through rigorous experimentation with these models, we have discovered that the sentiment of topics significantly enhances the performance of deep learning models in predicting stock market trends.

This discovery is particularly striking, as it reveals that the topics discussed in stock market comments contain implicit yet valuable information about stock market volatility and price trends. The integration of this sentiment analysis with deep learning models has proven to be a powerful combination, offering a more nuanced understanding of the factors influencing stock prices. As a result, this study not only contributes to the existing body of knowledge in financial analytics but also opens new pathways for future research. Such research could explore the real-time application of sentiment analysis in stock markets and delve deeper into the emotional and contextual aspects of market sentiment to yield even more refined insights.

In conclusion, the insights gained from this study contribute to the evolving field of financial analysis by integrating NLP techniques, particularly BERTopic, into stock price prediction models. Our findings suggest that this approach can enhance the understanding of market sentiment, potentially improving the accuracy of stock market predictions. While this represents a promising direction, it also highlights the need for ongoing research to further refine and validate these methods. Ultimately, this study serves as a stepping stone towards more advanced and nuanced financial analysis tools that can adeptly navigate the complex interplay of language and market dynamics.

\bibliography{ref}
\bibliographystyle{IEEEtran}
\end{document}